\documentclass[10pt,letterpaper]{article}
\usepackage[top=0.85in,left=2.75in,footskip=0.75in]{geometry}
\usepackage{booktabs}
\usepackage{amsmath,amssymb}

\usepackage{changepage}

\usepackage[utf8x]{inputenc}

\usepackage{textcomp,marvosym}

\usepackage{cite}

\usepackage{changes}

\usepackage{nameref,hyperref}

\usepackage[right]{lineno}

\usepackage{microtype}
\DisableLigatures[f]{encoding = *, family = * }

\usepackage{array}

\usepackage{appendix}

\newcolumntype{+}{!{\vrule width 2pt}}

\newlength\savedwidth

\raggedright
\usepackage[aboveskip=1pt,labelfont=bf,labelsep=period,justification=raggedright,singlelinecheck=off]{caption}

\makeatletter
\renewcommand{\@biblabel}[1]{\quad#1.}
\makeatother

\usepackage{lastpage,fancyhdr,graphicx}
\usepackage{epstopdf}
\fancyheadoffset[L]{2.25in}
\fancyfootoffset[L]{2.25in}
\begin{document}
\vspace*{0.2in}

\begin{flushleft}
{\Large
\textbf\newline{Neighbors and relatives: How do speech embeddings reflect linguistic connections across the world?} 
}
\newline
\\
Tuukka Törö*,
Antti Suni,
Juraj Šimko,
\\
\bigskip
\text Department of Digital Humanities, University of Helsinki, Helsinki, Finland

\bigskip

* tuukka.toro@helsinki.fi

\end{flushleft}
\section*{Abstract}

Investigating linguistic relationships on a global scale requires analyzing diverse features such as syntax, phonology and prosody, which evolve at varying rates influenced by internal diversification, language contact, and sociolinguistic factors. Recent advances in machine learning (ML) offer complementary alternatives to traditional historical and typological approaches. Instead of relying on expert labor in analyzing specific linguistic features, these new methods enable the exploration of linguistic variation through embeddings derived directly from speech, opening new avenues for large-scale, data-driven analyses.

This study employs embeddings from the fine-tuned XLS-R self-supervised language identification model \texttt{voxlingua107-xls-r-300m-wav2vec}, to analyze relationships between 106 world languages based on speech recordings. Using linear discriminant analysis (LDA), language embeddings are clustered and compared with genealogical, lexical, and geographical distances. The results demonstrate that embedding-based distances align closely with traditional measures, effectively capturing both global and local typological patterns. Challenges in visualizing relationships, particularly with hierarchical clustering and network-based methods, highlight the dynamic nature of language change.

The findings show potential for scalable analyses of language variation based on speech embeddings, providing new perspectives on relationships among languages. By addressing methodological considerations such as corpus size and latent space dimensionality, this approach opens avenues for studying low-resource languages and bridging macro- and micro-level linguistic variation. Future work aims to extend these methods to underrepresented languages and integrate sociolinguistic variation for a more comprehensive understanding of linguistic diversity.

\section*{Author summary}


\section*{Introduction}

Understanding and quantifying relationships among languages is a complex task, as languages differ across multiple dimensions, including phonology, syntax, morphology, lexicon, and prosody~\cite{evans2009myth}. These features evolve at varying rates and are shaped by diverse influences such as internal diversification, contact with other languages, and broader sociolinguistic factors.

Traditional approaches to analyzing linguistic variation typically rely on genetic and typological frameworks. Genetic approaches trace languages to shared ancestors and reconstruct proto-languages combining various sources of information such as linguistics, archaeology and human genetics, constructing assumed historical migration patterns of speaker communities~\cite{campbell2013historical}, while typological approaches classify languages based on structural similarities and differences~\cite{comrie1989language}. Both rely on predefined assumptions about historical events, cross-linguistic categories or comparative concepts~\cite{croft2002typology,haspelmath2010comparative}, which can introduce interpretive biases. Furthermore, these approaches demand significant manual effort, and their dependence on human-curated linguistic labels limits their scalability.

Languages are not static systems; they are dynamic, constantly evolving through processes of internal diversification and external influences. Variation manifests across multiple linguistic and typological dimensions, each evolving at different rates depending on cultural, social, and ecological contexts~\cite{evans2009myth}. For instance, internal diversification occurs when languages diverge from their parent forms, potentially giving rise to entirely new languages. This process is often complemented by external contact with other languages, leading to linguistic change through borrowing, convergence, or the creation of entirely new language systems, such as creoles and pidgins.

Traditional genetic approaches in historical linguistics typically represent language evolution using the family tree model, borrowing from biology~\cite{franccois2015trees}. This model conceptualizes languages as branching from a common proto-language spoken by an ancestral community, which eventually diverged due to geographic and social separation. While the tree model simplifies the complex web of interactions that characterize linguistic evolution, e.g., different rates of change of different linguistic features, it provides an easily accessible method to visualize global inter-relatedness among multiple languages and their groupings into families based on the analyzed set of features. 

To address the inherent limitations of family trees, alternative frameworks such as the wave model and network-based approaches have been developed.
For instance, the NeighborNet method constructs phylogenetic networks instead of trees, allowing for intersecting connections that reflect the multifaceted influences shaping languages~\cite{bryant2004neighbor}. These models move beyond the linear, bifurcating narratives of traditional approaches to embrace the complex, intersecting relationships that define language evolution.

In recent years, machine learning (ML) methods have offered new avenues for exploring language relationships, primarily by analyzing vast corpora of \textit{textual} material~\cite{bjerva2019language, oncevay2020bridging, ostling2017continuous}. By leveraging data-driven techniques, researchers can analyze linguistic patterns across large datasets without relying exclusively on predefined linguistic categories or historical assumptions. Such approaches not only complement traditional methods but also reveal new perspectives on the interplay of structure, history, and function in the evolution of human language.

Text based ML approaches derived from, e.g., language identification (LID) or machine translation systems, use embeddings from self-supervised deep learning models for analyzing linguistic variation without reliance on predefined theoretical frameworks. The general assumption of this kind of ML assisted analyses--shared by the present account--is that the topological relationships among the numeric latent representations of linguistic material (or of the languages themselves) systematically reflect the underlying typological relations among the analyzed languages. One of the issues with comparing world languages based on the textual material is the necessary step of coordinating different orthographies, and either converting the texts to a shared (potentially phonetic) script or use embeddings that might include orthographic rather than linguistic information. This is not the case when dealing with spoken language directly rather than its very lossy and arbitrary transformation to a textual form. Additionally, many languages of interest lack a written form, meaning speech is the only viable medium for their analysis. Speech captures the natural variability and language change more effectively than text, offering richer insights into linguistic dynamics. Speech also provides a universal signal, making it feasible to build massively multilingual models without being limited by the orthographic constraints of individual languages~\cite{babu2022xls, mms}. These models can then be fine-tuned to specific tasks, such as speech recognition, or language identification in our case.

In this paper we thus analyze the latent space embeddings of a fine-tuned self-supervised language identification (LID) model derived from purely \textit{speech} material of over 100 languages. Specifically, we use embeddings from the XLS-R model, fine-tuned on the Voxlingua107 corpus~\cite{valk2021slt}, to cluster languages based on their sound characteristics, such as phonetic, phonotactic and prosodic features implicitly present in the recordings. Using a corpus of recently collected spoken material, the method takes a synchronic view of languages, analyzing them as they are currently spoken, sidestepping many of the constraints and assumptions of traditional methods.

This paper shares some reasoning--albeit being methodologically very different--with the experimental work presented in Skirgaard \textit{etal.}~\cite{skirgaard2017some} where participants listened to utterances in various languages and were asked to guess the language in a forced choice test, and their choice patterns were used to quantify similarities and differences among the analyzed languages.

Our goal is to evaluate the extent to which the speech-based embeddings reflect known genealogical and geographical clusters and to identify patterns of linguistic variation that emerge from the model's latent space. This analysis builds on previous work that has used embeddings for studying dialectal variation~\cite{suni2019comparative,hiovain2022comparative,toro2024sociolinguistic,toro24_speechprosody} and language family relationships~\cite{hsieh2024self}, extending it to a broader typological context. The current approach is also inspired by the Voxlingua107 datasets' authors' observations using an x-vector model that language embeddings can reflect language family relations~\cite{valk2021slt}. 

We investigate the feasibility of using state-of-the art LID model embeddings for the scientific analysis of language relationships. We extend previous work by analyzing languages and data unseen in the LID fine-tuning. We compare the distances among languages derived from sound-based embeddings with other types of distances, such as lexical distance and geographical distances between the areas where the languages are primarily spoken, and analyze the interplay between the influence of geography and genealogy on how languages sound. In addition, we evaluate the robustness of the embeddings with respect to data availability, and explore which analyses best reflect relationships at local and global levels. We explicitly address several methodological issues such as the dimensionality of the latent space used for analysis and the size of the corpus necessary to derive replicable results.

By clustering languages based on the speech-based embeddings, we aim to provide new insights into the linguistic diversity and relationships among the world's languages, complementary to established typological methodologies. We will argue that the cases where our language-grouping results deviate from genealogy are interpretable consequences of multilingualism and geopolitical influences on the global and local soundscape of world languages.

\section*{Materials and methods}

\subsection*{LID model}

We used a state-of-the-art \texttt{voxlingua107-xls-r-300m-wav2vec}~\cite{alumae2022pretraining} LID model fine-tuned with the Voxlingua107 dataset~\cite{valk2021slt} on the 300M XLS-R foundation model~\cite{babu2022xls}.
The foundational model XLS-R is a pretrained self-supervised model based on wav2vec 2.0 architecture, trained with 128 languages with a total of 436k hours of speech data. The foundation model provides contextual speech representations directly from raw waveforms using a convolutional feature encoder and a Transformer architecture. The model was trained by masking portions of the audio input and predicting the correct latent representations. The language identification fine-tuning dataset consists of 6.6k hours of speech extracted from YouTube videos for 107 languages. The LID model pools the frame-level outputs of the XLS-R model to form utterance level representations, and adds a language classification head with one additional linear layer on top of the pooling layer. 
We chose to extract the embeddings from this final 512-dimensional layer, which is more directly tied to the language classification task.

\subsection*{Dataset}

As the source material for speech recordings, we used the test set of the Mozilla Common Voice 16.1 (CV)~\cite{ardila2020common}, a corpus containing read speech in 120 languages. We did not use the training set since it was used as pretraining material for the XLS-R model. Importantly, CV was not used for the LID model fine-tuning. The CV corpus features a diverse set of languages across various language families, but with an obvious bias towards large European languages. African language families are heavily under-represented, and Austronesian and Indigenous American language families are almost absent.

For the languages represented in the corpus, there are varying numbers of speakers reading aloud short, isolated sentences. The recordings were submitted by voluntary speakers, using their own recording equipment. The language skills and recording quality are not controlled, and vary widely. The audio files were loudness normalized to -23 lufs and converted to WAV files with a 16~kHz frame rate. 

We kept the recordings from those languages for which we could match the label assigned in CV with an ISO639 code and geographical coordinates from Glottolog~\cite{glottolog5.1}, and that contained at least 20 recordings in the test set. This way the number of the analyzed languages reduced to 106 languages; these languages are listed in Appendix. 

For the recordings from these 106 languages, we then extracted language recognition hypotheses as well as the 512-dimensional latent space embeddings from the \texttt{voxlingua107-xls-r-300m-wav2vec} LID model. 

Out of the 106 languages, 74 were also used for training the LID model. The recognition accuracy of the LID model for these languages on the CV material was 0.88, comparatively modest compared with the reported accuracy of 0.95 on the LID model development set. This drop reflects the general quality and within-language variability of the CV recordings. As we model each language by their average speaker in this work, it is preferable to exclude non-native speakers and very noisy recordings from the analyzed dataset. 
The outlier exclusion was performed by computing the Silhouette score for the 512-dimensional embeddings of each recording from all 106 languages. The higher Silhouette score the more closely is the given recording associated with the other recordings from the same language, i.e., the more reliable representative of the given language. 

Examining the scores for different languages, we identified two challenging sources of intralinguistic variation. On the one hand, for the globally spoken languages such as English, a variety of valid accents make it difficult to determine outliers (or, for the matter, to determine the average English speaker). On the other hand, for minority languages such as Celtic and some Uralic languages, a significant proportion of its speakers sound heavily influenced by the majority dominant language, forming bimodal distribution within those languages. Aware of these difficulties, we opted to treat all languages equally by simply removing 10\% of the lowest scoring recordings for each language. After the outlier removal, recognition accuracy rose to 0.92. 

We then acquired corresponding geographical coordinates and language family membership for each language in order to analyze the relationship between the embeddings, geography and genealogy. As a majority of represented languages are Indo-European, we further split this family to Classical Indo-European subfamilies according to the Glottolog hierarchy.  Finally, we randomly subsampled the data to a maximum of 1000 recordings per language (actual numbers per language listed in the appendix).

\subsection*{Language embeddings and distances}

\begin{figure}[!ht]
\caption{{First five PCA and LDA components of the Common Voice utterance embeddings}}
\label{fig:Fig1}
\includegraphics[width=1.0\linewidth]{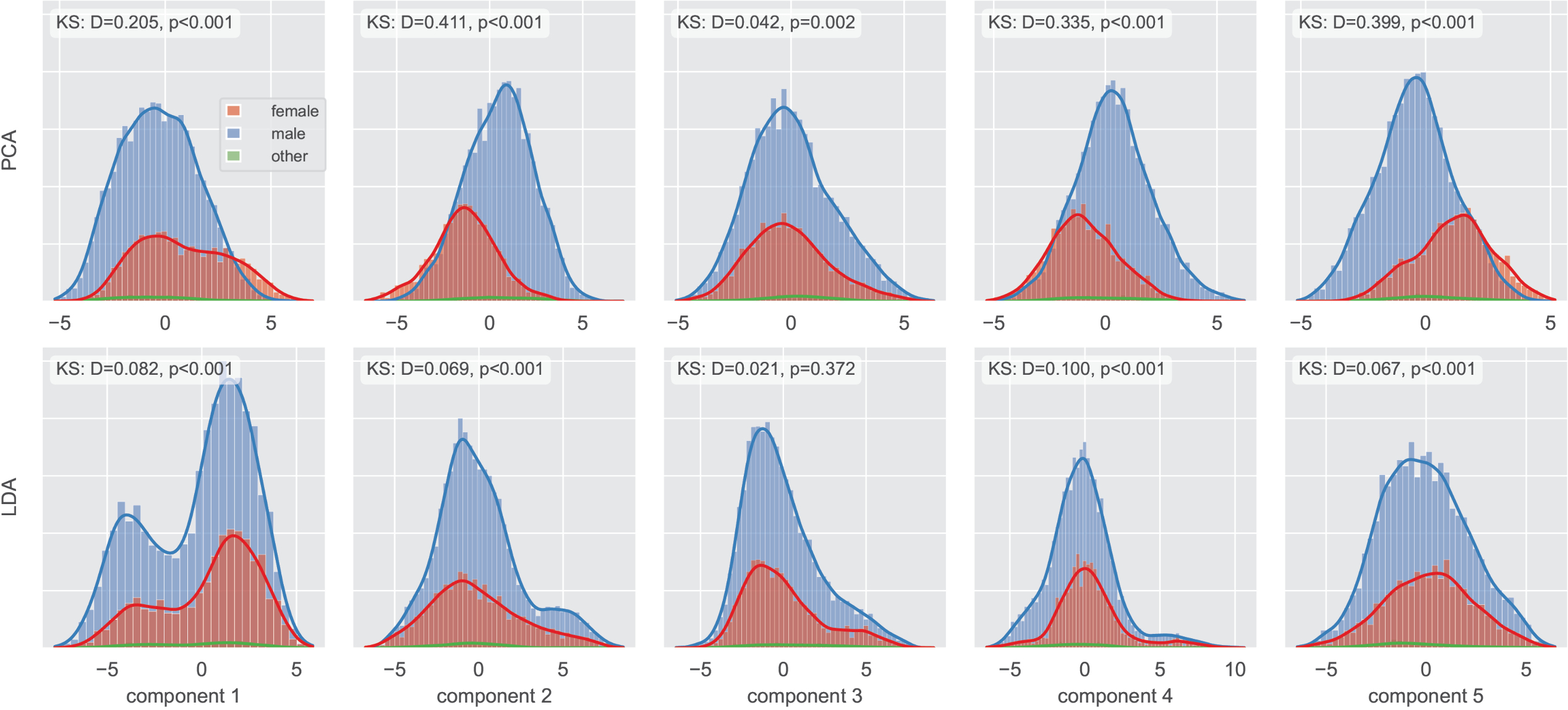}
\end{figure}
Fig~\ref{fig:Fig1} (top row) displays histograms of the first five PCA components derived from 512-dimensional embeddings from the LID model, for the utterances in the CV-based dataset described above. These components, accounting for 18.8 \% of total variance, visibly separate male and female utterances. Two-sample Kolmogorov-Smirnov (KS) tests confirm these distributional differences for four of the five components (see annotations in Fig.~\ref{fig:Fig1}). This suggests the LID model may utilize gender-correlated features, potentially stemming from varying gender distributions of languages within the training data, a concern also discussed by the VoxLingua authors regarding potential biases\cite{valk2021slt}.

\begin{figure}[!ht]
\begin{adjustwidth}{-2.25in}{0in} 
\caption{{\bf Languages in the linear discriminant space comprised of the first two discriminants.} Already, the first two LDs project the high-level relationships quite well. The longitudinal West-East continuum of European and Central Asian languages are represented on the 1st LD, and a split between European and non-European languages on the 2nd LD. For fine-grained relationships, more LDs are needed.}
\label{fig:Fig2}
\includegraphics[width=1.\linewidth]{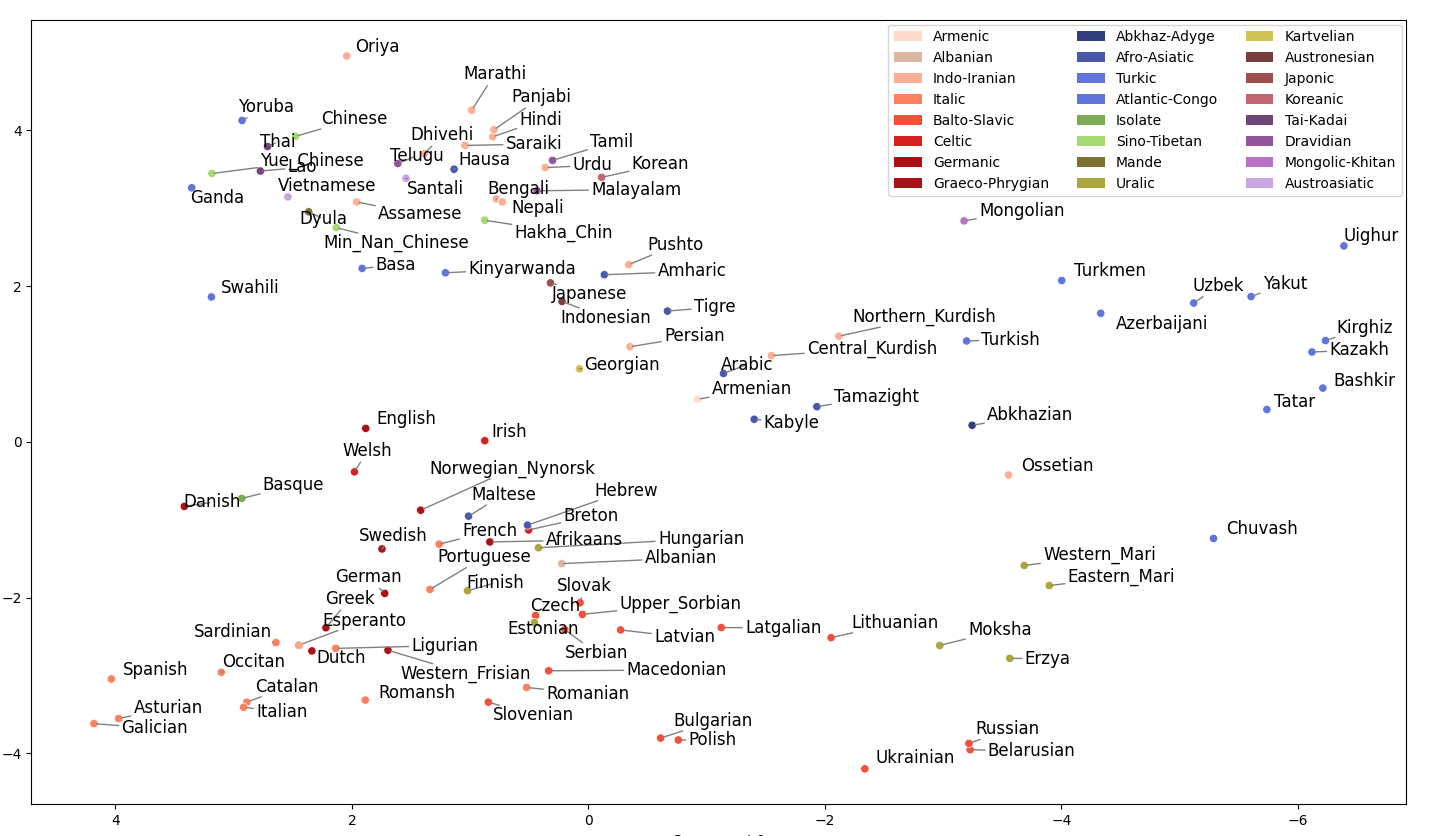}
\end{adjustwidth}
\end{figure}

To reduce this type of within-language variance and highlight between-language differences in the language representation, we fitted a linear discriminant analysis (LDA) model on the embeddings using language labels as classes. The dimensionality of the LDA-transformed space is limited by the number of classes, i.e., languages (being one less that the number of classes). As shown in the bottom row of histograms in Fig~\ref{fig:Fig1}, while the first five LDA components are still statistically influenced by the gender information (presumably due to the very large number of data points), as indicated by the considerably lower distribution distance values $D$, the relationship is considerably weaker than for the PCA components. Finally, we used the centroid of the language samples in the LDA space as a \textit{language embedding}, i.e., the representation of the given language in subsequent analyses. Fig~\ref{fig:Fig2} shows the first two dimensions of this space with the appropriate language labels colored by the language families.

Subsequently, we used cosine distances between the L2-normalized (using the SciPy Python library~\cite{2020SciPy-NMeth}) multi-dimensional LDA-based language embeddings as a proxy for the distances among the analyzed languages.

In order to evaluate how the \textit{embedding distances} reflect genealogical relationships between the languages, we used \textit{lexical distances} derived from ASJP database~\cite{wichmann2022ASJP} as an approximation of genealogical distances. ASJP contains a comparable list of 40 core vocabulary words in the world's languages, and the language distances are calculated by comparing the phonological transcriptions by the Levenshtein Distance Normalized Divided method introduced in \cite{bakker2009adding} that is based on the normalized number of edits needed to make the transcriptions identical. Unlike genealogical distance measures that are based on family trees, the lexical distance provides a naturally continuous measure, and it is not affected by varying granularity between different language family trees. The ASJP based Levenshtein distances have been previously shown to robustly correlate with established language family trees~\cite{pompei2011accuracy}. The ASJP database does not contain word lists for all of the 106 languages in our dataset, and therefore the lexical distance was only calculated for 101 of them. This somewhat limits the scope of the analysis presented below and in Fig~\ref{fig:Fig3}. Finally, for each pair of languages we calculated geographical distances based on the coordinates provided in Glottolog.

\subsection*{Visualization}

We used three different approaches to visualize relationships between languages in the LDA-space. We plotted the languages on a world map, drawing lines between languages based on the nearest 5th percentile between language pairs in the space. For clustering the languages we used both Hierarchical Agglomerative Clustering (HAC) and the NeighborNet algorithm. For all analyses, we used cosine distance as our distance metric.

HAC works bottom-up by clustering together leaves and clusters that are nearest, finally connecting all leaves into a complete tree. We used the UPGMA algorithm for clustering which takes the distance between the average of the leaves in each cluster to determine the linkages~\cite{2020SciPy-NMeth}. The NeighborNet algorithm also works bottom-up, but unlike HAC it can create overlapping clusters. Instead of a tree, it constructs a phylogenetic network that can take into account intersecting connections between languages that more realistically represent the complex nature of language evolution.~\cite{franccois2015trees}

\section*{Results}

\subsection*{Correlations with geography and lexicon based distances}

\begin{figure}[!ht]
\caption{{\bf Correlations between embedding distances and other distance measures.}
}
\label{fig:Fig3}
\includegraphics[width=0.8\linewidth]{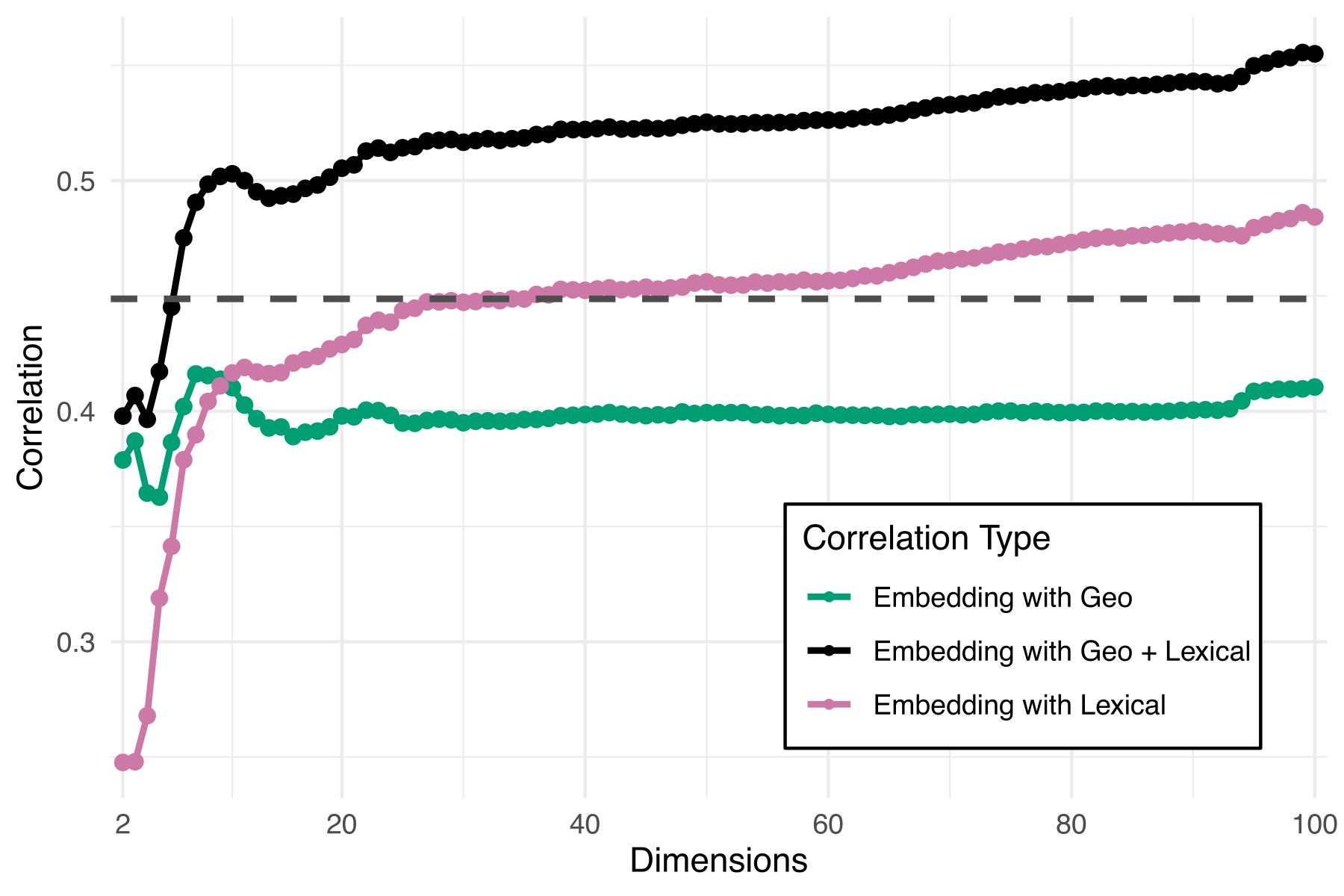}
\end{figure}

The LDA-based language embeddings are highly multidimensional representations of the shared sound characteristics of the languages, and at the outset it is not clear whether, in particular, higher dimensions of the latent space contribute to depicting relevant similarities and differences among languages. For example, the first two dimensions of this space plotted in Fig~\ref{fig:Fig2} clearly capture rudimentary geographical relationships between the languages, separating the European languages (in the bottom left) from the remaining languages in the database.
To verify the degree to which cumulative dimensions of the LDA transformed embedding space reflect both geographical and lexical relationships, we calculated the embedding distances using only the first $n$ LDA dimensions, with $n$ progressively increasing from $2$ to $100$ (the actual dimensionality of the LDA space for 101 languages). The green and magenta lines in Fig~\ref{fig:Fig3} depict Pearson's product-moment correlations between these LDA-based embedding distances, and geographic and lexical distances among languages. Interestingly, the correlations with geography-based distances peak for a smaller number of dimensions ($ < 10$), confirming that, as suggested above, the few first LDA embeddings delimit the broad sound-based characteristics of languages spoken in different regions. The correlations with lexical distances steadily increase with added LDA dimensions, indicating that even the higher dimensions of the LDA space contain information reflecting genetic relatedness among the analyzed languages.

For each $n$ we also fitted a linear model with cosine distances calculated from the first $n$ LDA dimensions as a dependent variable, and the geographic and lexical distances as independent variables (with interactions). In order to correct for skewness of the distributions, we used exponential transformation for embedding and lexical distances, and square-root transformation for geographical distance variable. The black line in Fig~\ref{fig:Fig3} represents the square roots of the adjusted $R$-square measure for these models, i.e., the correlation between the predictions of the model and the actual LDA-based distances. The fact that these correlations are greater than correlations with the individual measures shows that the LDA-transformed embeddings combine both geography and lexicon (genetics) based influences on the sound patterns of the analyzed languages. Also, the correlation (quality of fit) increases with added dimensions; therefore in the remainder of this paper we use the embedding-based distances calculated using all available dimensions of the LDA space. For comparison, the gray dashed line shows the square root of the adjusted $R$-square measure of the corresponding model with cosine distances calculated directly from the 512-dimensional embeddings; the LDA-based distances yield considerably better fits.

\begin{figure}[!ht]
\begin{adjustwidth}{-2.25in}{0in} 
\caption{{\bf Correlations between embedding distances and other distance measures.}
Cosine distance on a 100 dimensional LDA space.}
\label{fig:Fig4}
\includegraphics[width=1.\linewidth]{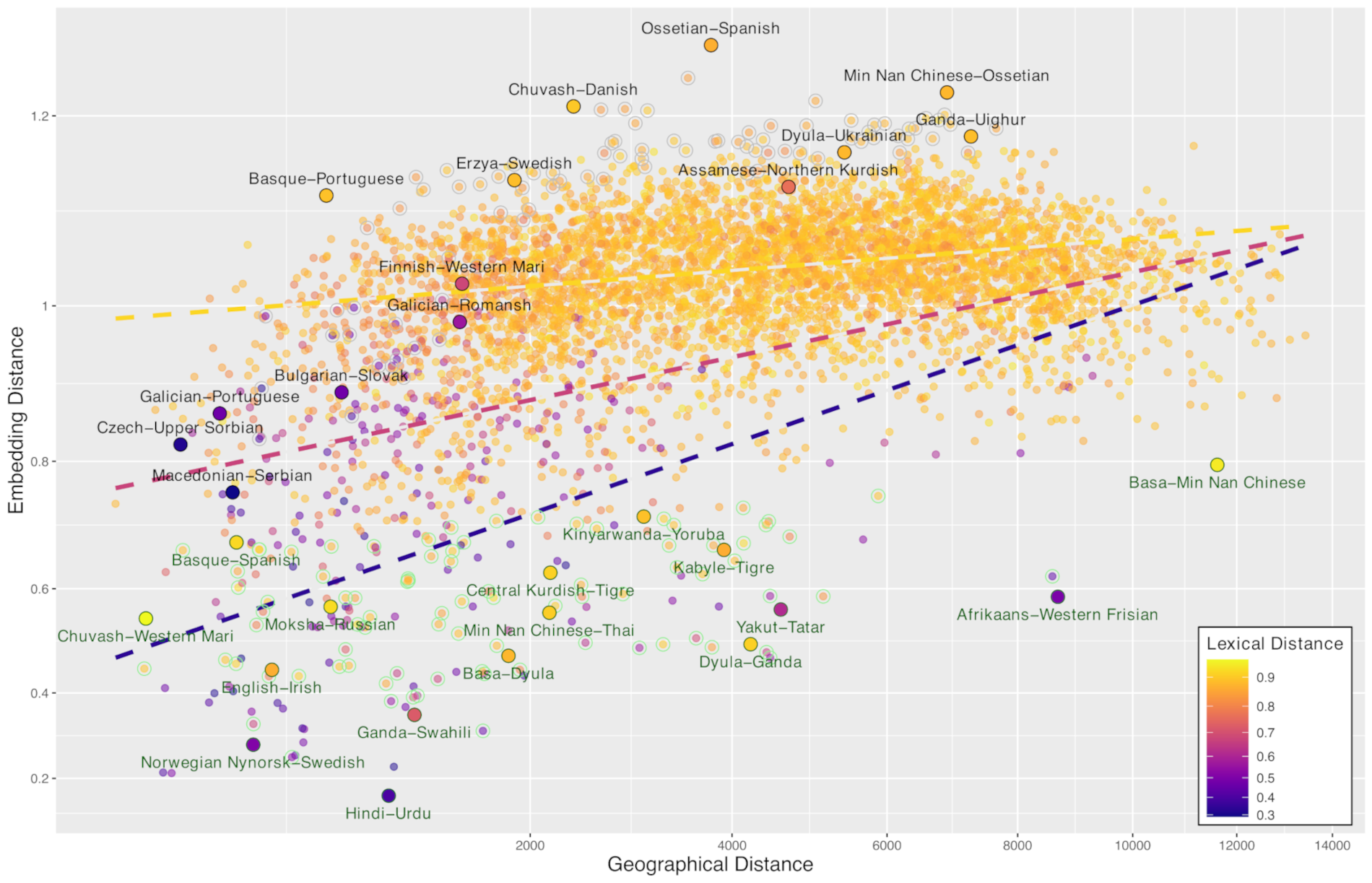}
\end{adjustwidth}
\end{figure}

The adjusted $R$-square value of the ``full'' linear model with embedding distances calculated using all LDA dimensions is $0.31$, corresponding to a correlation equal to $0.56$. Both main effects of geographic and lexical distances are significantly positive ($t$-values 9.66 and 21.74, respectively, $p<0.001$ for both). The interaction is significantly negative ($t = -8.16$, $p < 0.001$). Fig~\ref{fig:Fig4} plots the embedding distances between language representations used in the full model against geographic distances between the languages; lexical distance is depicted by the point color. The dashed lines correspond to projections of the fit to the embedding-geographical distance plane for lexical distance values fixed to $1.40, 1.95$ and $2.50$. Given the significantly negative value of interaction of the fit, the slope of influence of geographical distance on the embedding distance steadily decreases--from blue to yellow dotted line--with increasing lexical distance (i.e., for less genetically related languages).

The language pairs with the greatest absolute residual from the fit--i.e., the outliers regarding the relationship between geographical-lexical distances and embedding distance--are highlighted in Fig~\ref{fig:Fig4}; 100 pairs with greatest positive residual (greater embedding distance than predicted by lexicon and geography) by gray circles, 100 pairs with the greatest negative residual in green (closer embedding distance than expected). Some of the highlighted outliers are labeled in the figure; the fill color of the corresponding data-point corresponds to the coloring of the dashed line along which the regression model predicts its position based on the geographical and lexical distances.

For some closely related languages (plotted in shades of magenta), the fit predicts a greater embedding distance than the actual one, because they are geographically more distant than suggested by their genetic similarity; examples being Afrikaans and Western Frisian but also Nynorsk--Swedish and Hindi--Urdu. The last pair highlights one of the limitations of calculating the geographical distance using fixed `central' coordinates for a language: Hindi and Urdu are in fact spoken in overlapping geographical areas, and the embedding distance (smallest of all in our dataset) reflects their close geographic and genetic proximity.

For other pairs, the prediction goes the other way, for example the actual embedding distance is greater than expected for several pairs of Slavic languages (Macedonian--Serbian, Czech--Upper Sorbian, Bulgarian--Slovak) but also between Portuguese and fellow Italic Galician and Romansh, the Uralic pair Finnish--Western Mari and Indo-Iranian Assamese and Northern Kurdish.

Other outliers consist of genetically distinct language pairs (plotted in shades of yellow). For some of them, the embedding distance is again smaller than predicted by the fit, in most cases presumably because the languages are spoken in vicinity of each other, often conditioned by majority language influence and multilingualism. Examples of these are Turkic Chuvash and Uralic Western Mari, Slavic Russian and Uralic Moksha, all languages spoken in Russia, but also Celtic Irish and Germanic English spoken in British Isles, Isolate Basque and Spanish spoken on the Iberian peninsula, and Mande Dyula and Atlantic-Congo Ganda spoken in relatively distinct parts of Africa. On the end of spectrum, the embedding distance between Basque and Italic Portuguese is greater than predicted by their geographical and lexical distances, and several other pairs plotted on the top part of Fig~\ref{fig:Fig4}.

An interesting case of a language pair with a smaller embedding distance than predicted by their geographical and lexical distance is Atlantic-Congo Basa spoken in Cameroon in Western Africa and Sino-Tibetan Min Man Chinese. Both of these tonal languages use a tonal system that can be broadly classified to high, low, rising and falling categories, and this tonality might elicit similar surface sound patterning between the languages.  

\begin{table}[!ht]
\begin{center}
\begin{tabular}{l|c|ccc|ccc}
\hline
    &    \#    & emb- & emb- & emb- & geog- \\
Family & langs & geog & lex  & geog+lex  & lex \\
\hline
\textbf{Afro-Asiatic} & 8 & 0.1 & 0.4 & 0.38 & 0.59 \\ 
\textbf{Turkic} & 11 & 0.16 & 0.47 & 0.44 & 0.16 \\ 
\textbf{Uralic} & 7 & 0.9 & 0.61 & 0.91 & 0.57 \\ 
\textbf{Indo-European} & 52 & 0.49 & 0.64 & 0.69 & 0.4 \\ 
\hline
Balto-Slavic & 13 & 0.26 & 0.55 & 0.55 & 0.28 \\ 
Germanic & 8 & 0.3 & 0.48 & 0.58 & -0.22 \\ 
\textit{Germanic w/o Afrikaans}  & 7 & 0.72 & 0.5 & 0.84 & 0.38 \\ 
Indo-Iranian & 15 & 0.82 & 0.63 & 0.83 & 0.62 \\ 
Italic & 10 & 0.37 & 0.5 & 0.55 & 0.17 \\ 

\hline
\textbf{Overall-related} &  & 0.47 & 0.66 & 0.7 & 0.39 \\ 
Overall-nonrelated &  & 0.31 & 0.06 & 0.32 & 0.08 \\ 
\hline
\textbf{Overall} & 102 & 0.41 & 0.48 & 0.56 & 0.36 \\ 
\end{tabular}
\caption{Correlations between different types of language distance measures, including correlations for selected language families and groups.}\label{tab:corrs}
\end{center}
\end{table}

Table~\ref{tab:corrs} lists correlations among the various measures of distances among languages introduced here (Pearson's product-moment correlations between individual distance measures and square root of adjusted $R$-square measure for respective linear model fits; see above). The overall measures listed in the bottom row correspond to the rightmost data points in Fig~\ref{fig:Fig3}. In addition to the overall measures we also present correlations for selected language group subsets (families with more than 5 languages in the analyzed database), as well as for distances among pairs of related languages (i.e., the languages belonging to the same family) and a complementary subset of distances among the unrelated languages. The last column shows respective correlations between the two reference distance measures.

All correlations in Table~\ref{tab:corrs} are positive, except for the correlation between geographical and lexical distance for the Germanic subfamily. This discrepancy can be attributed to the effect of Afrikaans, which is genetically relatively closely related to the other Germanic languages but geographically separated. As seen in the table, the remaining Germanic languages follow the same trend as other language (sub)groups with the strong positive correlations among the distance measures.

In the vast majority of cases (Afro-Asiatic group being one of the few exceptions), the correlations between the reference distance measures (lexical, geographical) and embedding distances is greater than the correlation between the reference measures themselves. In particular, the correlation with geographical distances is higher for the embedding distances than for the lexical ones. This does indicate a stronger influence of language contact on the way languages sound compared to their genetic relatedness. 
Apart from the geographically wide-spread Uralic family, embedding distance correlates better with lexical distance than the geographical one (this is, for obvious reasons, also not the case for the unrelated languages). The correlations are considerably higher for the subgroup of related languages than overall. The correlations for the Indo-European family dominating our data set are almost identical to those for related languages. Interestingly, the embedding distance correlates considerably better with the lexical than the geographical one for both Afro-Asiatic and Turkic families, even though, based on the correlations between geography and lexicon, the lexical similarities follow geographical distribution for the former group but not for the latter. The correlations are exceptionally high for the Uralic family, in particular the geography-based ones; this is presumably due to the fact that this family is represented by three geographically separated subgroups.

\subsection*{Visualization of the embedding distances among languages}

\begin{figure*}[!h]
\begin{adjustwidth}{-2.25in}{0in} 
\caption{{\bf A distance matrix of LDA-transformed language embeddings, sorted by language families.}}
\label{fig:Fig5}
\includegraphics[width=0.99\linewidth]{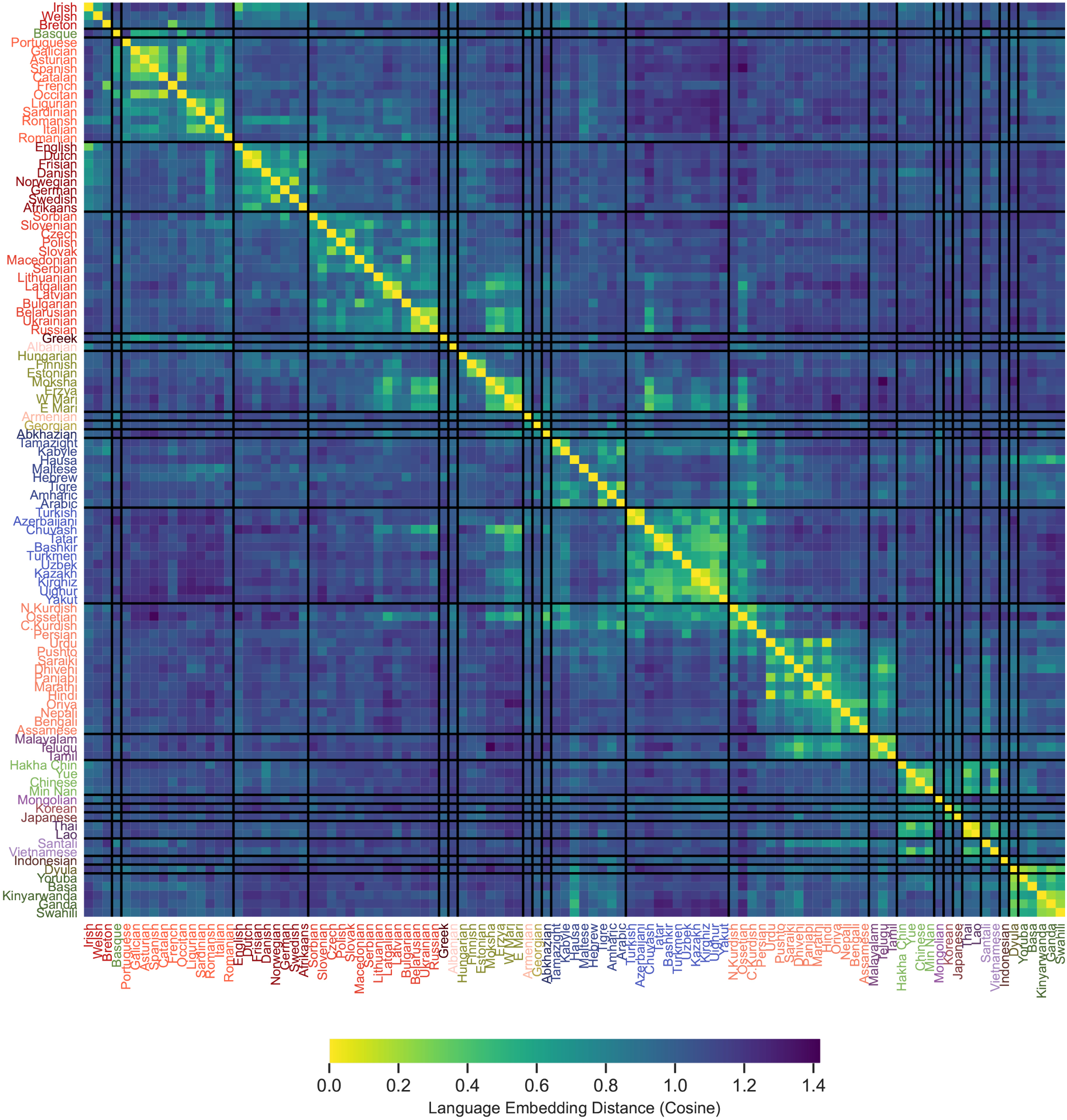}
\end{adjustwidth}
\end{figure*}

Fig~\ref{fig:Fig5} presents a summary of the embedding distances between the languages grouped by their (sub)families. Along the diagonal, the areas of smaller distances (in lighter green or yellow) clearly indicate close interrelatedness between languages from the same linguistic group. The low-distance areas off the diagonal show interesting connections between less related languages. For example, the Mande language Dyula embedding is close to the embeddings of Bantu languages, namely Basa and Yoruba spoken in the neighboring geographic areas. Dravidian languages show similarity with Indo-Iranian languages also spoken in India. Austroasiatic Vietnamese embedding is close to the Tai-Kadai Lao and Thai as well as to Yue Chinese spoken in Southern China across the Northern Vietnamese border. The embeddings of Uralic languages spoken in Russia--Ezra, Moksha and Mari languages--are closer to Eastern Slavic languages than to the Uralic ones from the Fenno-Ugric subfamily. They are also closer to a group of Turkic languages like Bashkir, Chuvash, Tatar and Yakut; this similarity might be more likely attributed to the influence of the majority Russian language than to a genetic relatedness.

\begin{figure*}[!h]
\begin{adjustwidth}{-2.25in}{0in} 
\caption{{\bf Average embedding distances of languages from $n$ nearest neighbors.}}
\label{fig:Fig6}
\includegraphics[width=0.99\linewidth]{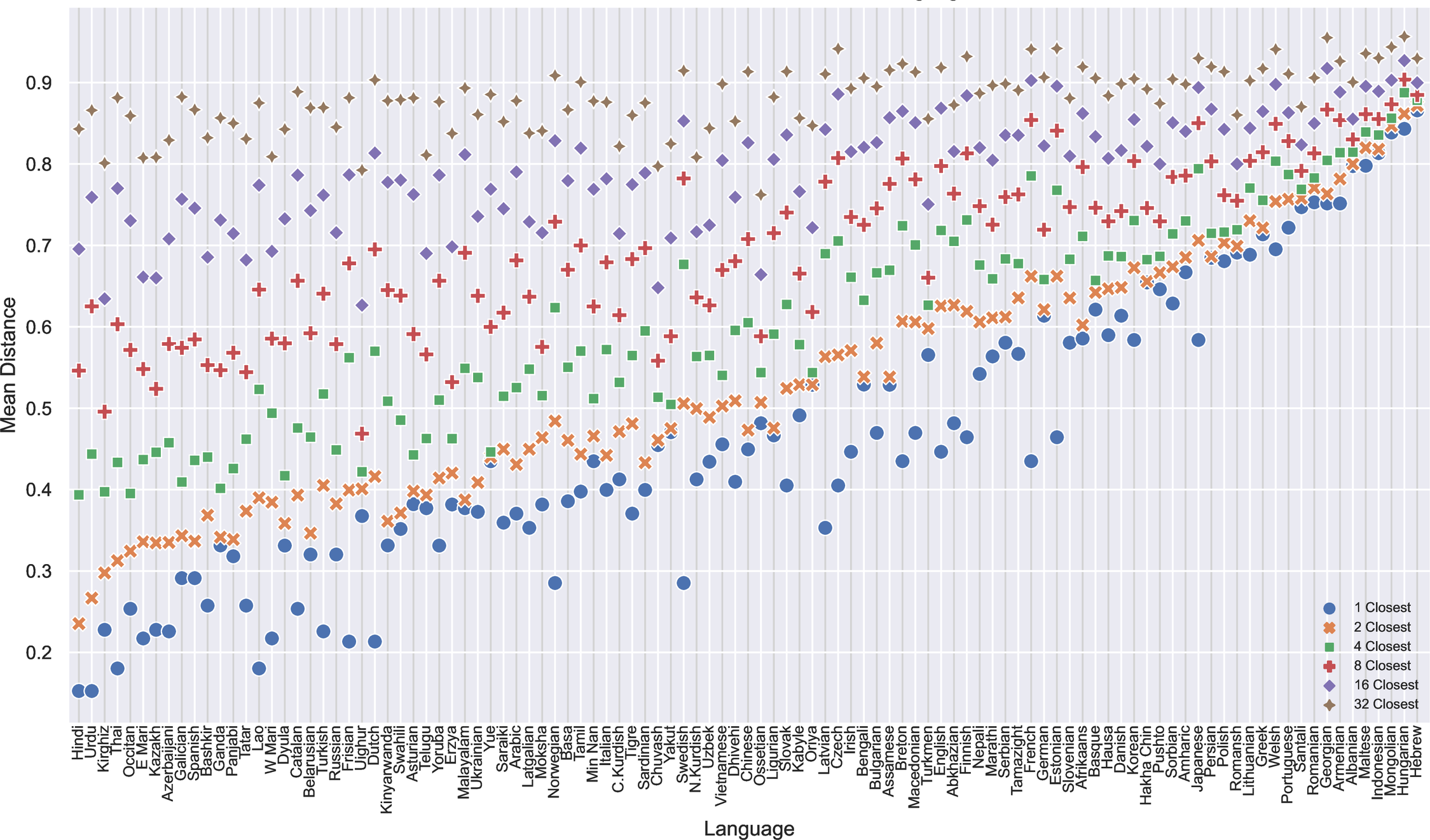}
\end{adjustwidth}
\end{figure*}

Fig~\ref{fig:Fig6} sorts languages by their distances from several nearest neighbors. While distances from the nearest neighbor vary among the languages to a large extent, the average distance from a larger group of neighbors (32 in the figure) does not show a clear trend. This implies that the primary source of the differences captured by the embedding space is the distance from a relatively small group of a few closest languages. The languages in the rightmost side of the figure are far from their nearest neighbors in our dataset and are approximately equidistant from their single nearest neighbor and a large group of closest languages. This group contains languages, such as Hebrew, Hungarian, Mongolian, Indonesian, Maltese, Armenian and Georgian, that are without genealogically close languages in the dataset.

\begin{figure}[!ht]
\begin{adjustwidth}{-2.25in}{0in} 
\caption{{\bf 5th percentile of closest pair-wise embedding distances between languages.}
Within language family connections represented by colored lines, and across family similarities with dashed white lines. Line weight is associated with the degree of embedding similarity. Languages without close connections are labeled. World map sourced from Nasa Blue Marble Collection~\cite{nasa_blue_marble}.}
\label{fig:Fig7}
\includegraphics[width=1.\linewidth]{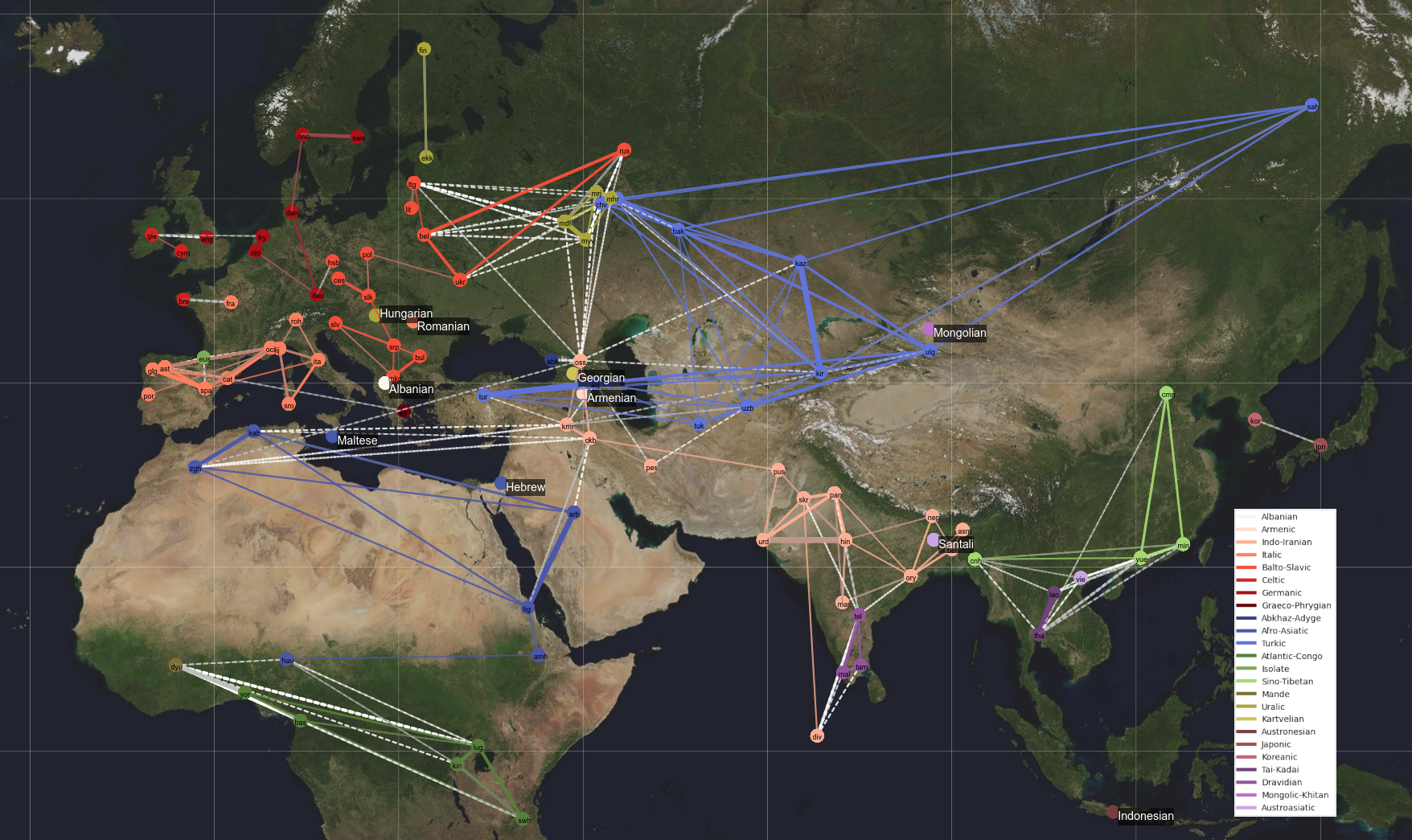}
\end{adjustwidth}
\end{figure}

Fig~\ref{fig:Fig7} shows a visualization of the embedding distances projected on the geographical map of the relevant world region. The lowest 5~\% of embedding distances are shown as lines joining the geopolitical centers of the respective language use, colored by language (sub)family (where both joined languages belong) or a white dotted line in the case of unrelated language pairs. 

The map shows close embedding distances relating to language families as well as to geography –– influenced by terrain. Especially in Europe, languages tend to be connected mostly within their subfamilies. This might be the result of a longer history of national borders separating language communities and a high-level of language standardization. There are no connections across the Sahara desert while North Africa and West Asia are closely connected. South Asian languages are in their own group (the only close connection being between Pashto and Central Kurdish). This could be due to mountain ranges such as the Hindu Kush and the Himalayas splitting South, Central and East Asia. Similarly, Georgian and Armenian spoken in the Caucasus are not close to any languages.

\begin{figure}[!h]
\caption{{\bf A single tree of 106 world languages}
based on cosine distance in a 105 dimensional LDA space. Languages not in the LID model training data are marked with an asterisk.}
\label{fig:Fig8}
\includegraphics[width=1.\linewidth]{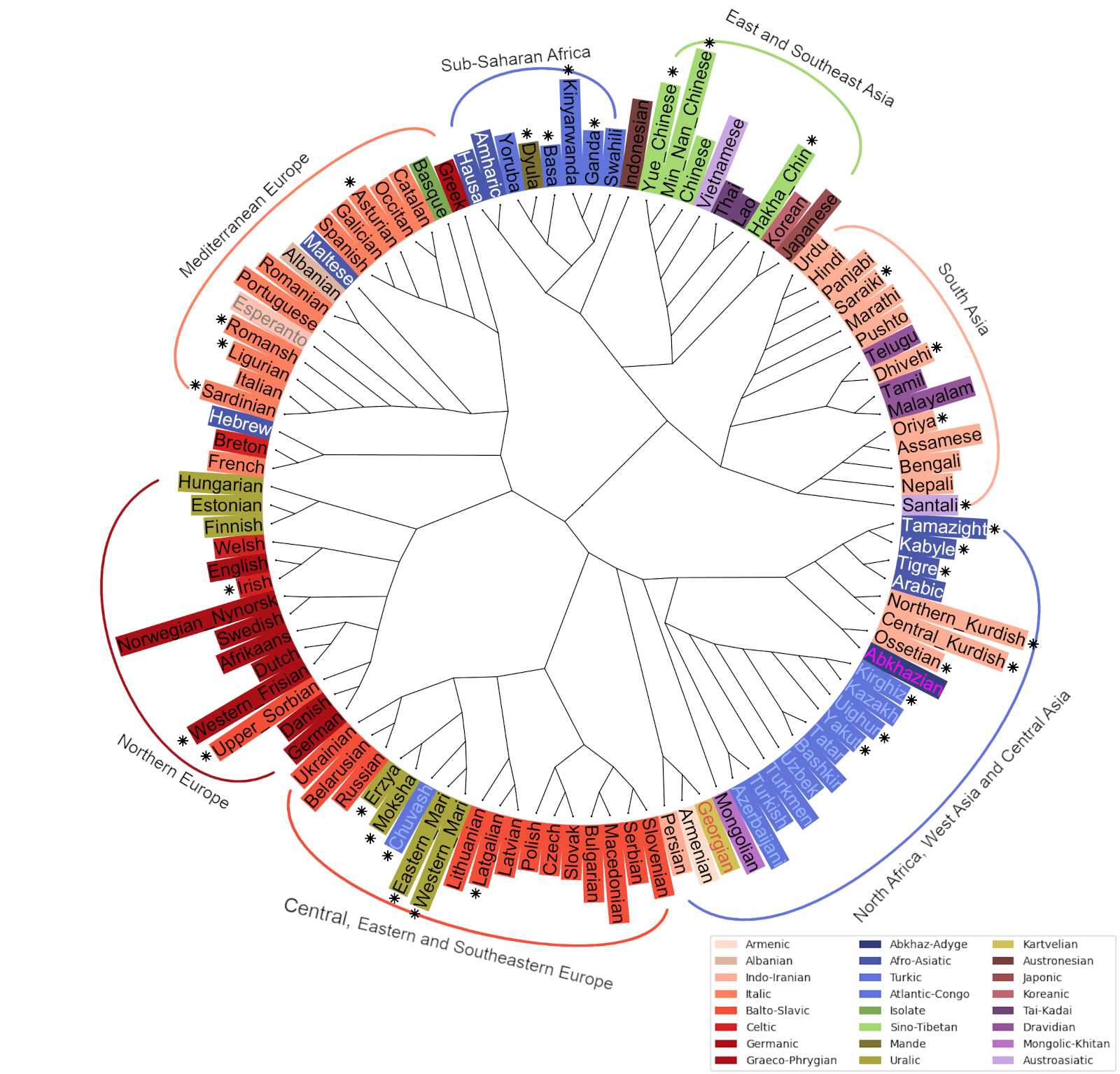}
\end{figure}

\begin{figure}[!h]
\caption{{\bf A NeighborNet network of 106 world languages}
 based on cosine distance in a 105 dimensional LDA space. Languages not in the LID model training data are marked with an asterisk. Created using SplitsPy Python library.~\cite{bagci2021microbial} }
\label{fig:Fig9}
\includegraphics[width=1.\linewidth]{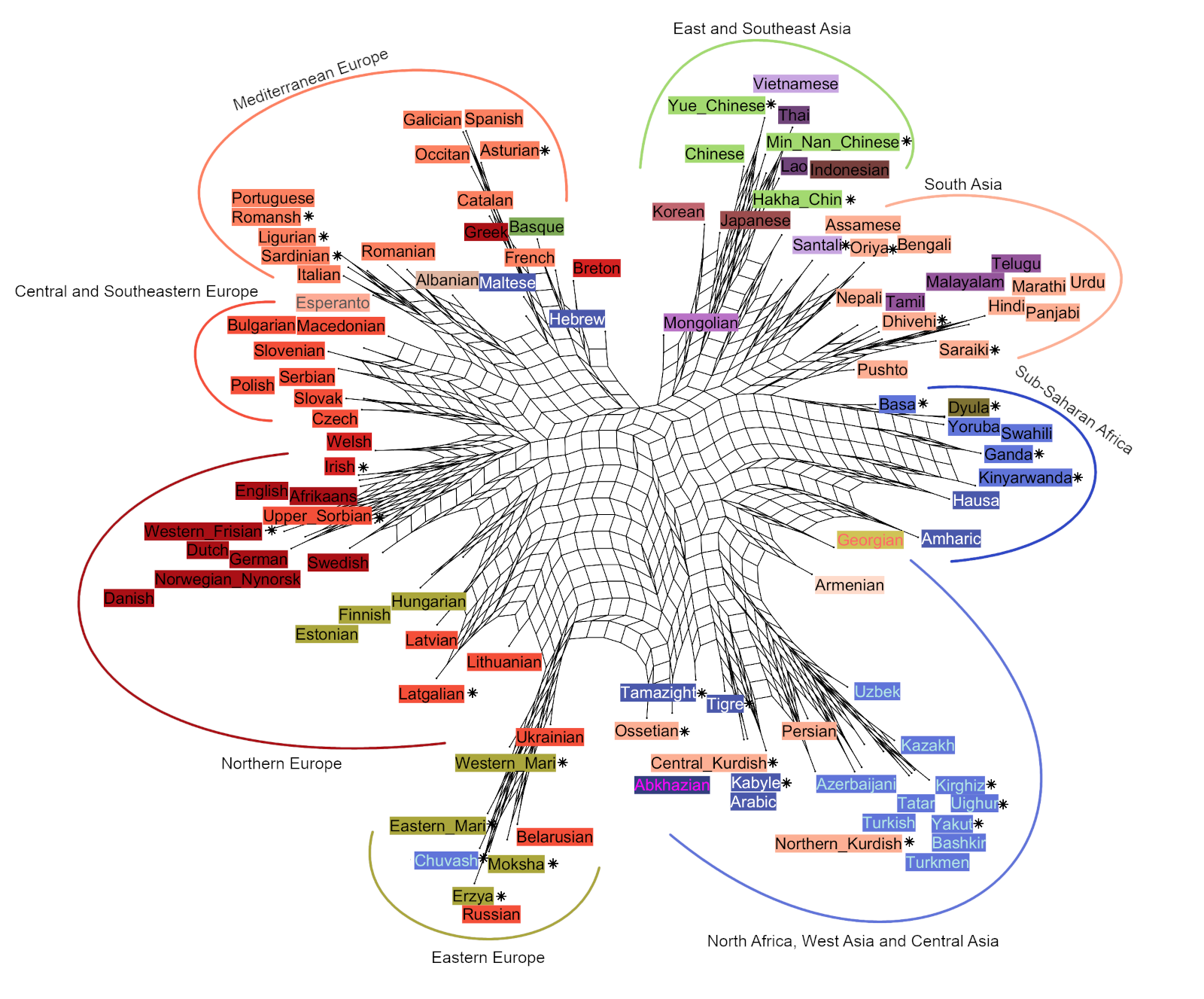}
\end{figure}

Figs~\ref{fig:Fig8} and \ref{fig:Fig9} represent the relationships among languages based on the LDA embedding space. The dendrogram in Fig~\ref{fig:Fig8}, created using HAC, reflects the traditional phylogenetic tree form of relatedness among languages, while the network-based NeighborNet visualization in Fig~\ref{fig:Fig9} represents the distances between languages as a web-like graph rather than a strict tree showing both tree-like and non-tree-like relationships.

The overall picture in Fig~\ref{fig:Fig8} largely follows expected genealogical and geographical relations. Sub-Saharan Africa, East Asia and South Asia cluster on one side while all the rest – European, North African and West and Central Asian languages – cluster on the other. Remarkably, apart from a few outliers such as Hebrew, Greek, and Indonesian (cf. Fig~\ref{fig:Fig6}), which seem to be placed on spurious similarities, connections in the tree reflect either family relation or language contact due to geographic proximity.

Sub-Saharan languages cluster together with three subgroups: Afro-Semitic languages, Bantu languages and Western African languages. Indonesian with no close genetic relatives in the dataset clusters with the group as an outlier. In the East and Southeast Asian cluster, Yue, Min Nan, Mandarin and Vietnamese cluster together after which Thai and Lao join the cluster together, then Hakha Chin and finally Korean and Japanese as a pair. The South Asian languages divide into two or three clusters. Languages spoken in Northern India and Pakistan cluster together first and then join with languages spoken mainly in Southern India. This North-South cluster is then joined together with languages spoken in the Northeast of the Indian subcontinent.

Slavic languages as well as those Uralic languages spoken within Russia cluster together with Turkic and other languages spoken in Central and West Asia and North Africa. Rest of the European languages cluster together with (mostly) Germanic languages in one large group and (mostly) Italic languages in the other. Incidentally, Greek being close to Spanish and Portuguese close to Romanian is consistent with \cite{skirgaard2017some} that used a language distance based on human subject judgments.
Within the Germanic group, there are languages from other subfamilies: Upper Sorbian, Irish, Finnish, Estonian and Hungarian. In the Italic group there are Basque, Greek, Maltese, Albanian, Hebrew and Breton. Most of these are clearly to do with a high-level of contact and multilingualism: Upper Sorbian with German and Irish with English. Breton is spoken in France and Basque mainly in Spain. Maltese is a Semitic language with a long history of contact with Italic languages while Finnish and Estonian are Uralic languages with contact with Germanic languages. In the case of Hungarian, language genetics seem to contribute more to its location in the embedding space than contact while the opposite is true with Erzya, Moksha, Eastern Mari and Western Mari that cluster with Slavic languages.

Balto-Slavic languages have an East-West split with Ukrainian, Belarusian, Russian, Lithuanian, Latvian, Latgalian and Uralic languages in one cluster and the rest in the other. Turkic languages form a clear group regardless of geography. They are joined by Semitic languages – Tamazight, Kabyle, Tigre and Arabic – that first cluster with Kurdish languages, as well as Ossetian and Abkhazian. Finally, the large cluster is joined by Mongolian, Georgian, Armenian and Persian before joining with the Balto-Slavic and Uralic cluster.

When we look at the NeighborNet network in Fig~\ref{fig:Fig9}, it is largely similar to the dendrogram, but there are some differences: On the tree, Mongolian is in the Central Asian cluster while in the NeighborNet it is next to Korean and Japanese. East and West Slavic languages have also split with West Slavic languages residing between Italic and Germanic languages. Estonian, Finnish and Hungarian are between Scandinavian and Baltic languages which, in the case of Finnish and Estonian, reflects geographical proximity. In the dendrogram, Central and Northern Kurdish both cluster with Semitic languages while in the network, Central Kurdish is near Arabic and Northern Kurdish near Turkish, reflecting influence from majority languages in the areas where the two languages are spoken. This highlights the possibility to represent relationships in the same data in multiple meaningful ways and how challenging it is to account for multidimensional influences between languages projected in two dimensions.

\subsection*{Robustness and data size}

\begin{figure}[!h]
\caption{{\bf Sensitivity of correlations among the cosine distances for individual languages as a function of sample size.} The mean and standard deviation across all languages in gray, and several outlier languages. The contours were smoothed using LOESS smoothing technique with span 0.2.}
\label{fig:Fig10}
\includegraphics[width=1.\linewidth]{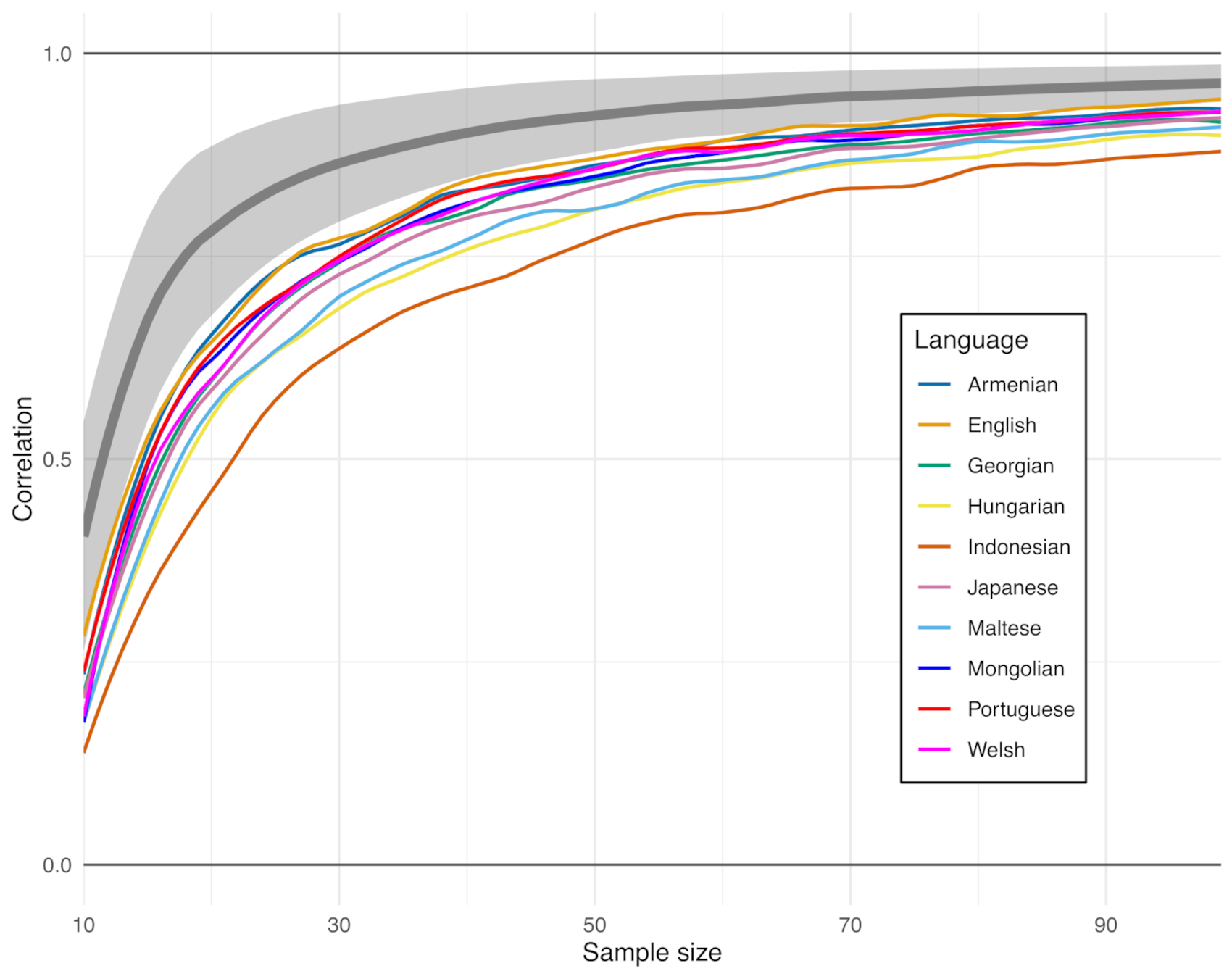}
\end{figure}

The relationships among the languages shown in Figs~\ref{fig:Fig8} and \ref{fig:Fig9} are based on the existing samples in our database, ranging between 20 and 1000 individual recordings. It is natural to ask, how does the sample size for an individual language influence both the overall distribution and local details of inter-language relationships obtained using the presented method. This question is closely linked to the one of a minimal size of a dataset (in terms of samples per language) that yields interpretable and robust results.

\begin{figure}[!h]
\caption{{\bf A consensus tree of 70 languages.} The tree summarizes commonalities among 20 individual trees, each with non-overlapping 50 samples of each of the 70 languages with at least 1000 samples available. We took cosine distance from a 69 dimensional LDA space.}
\label{fig:Fig11}
\includegraphics[width=1.\linewidth]{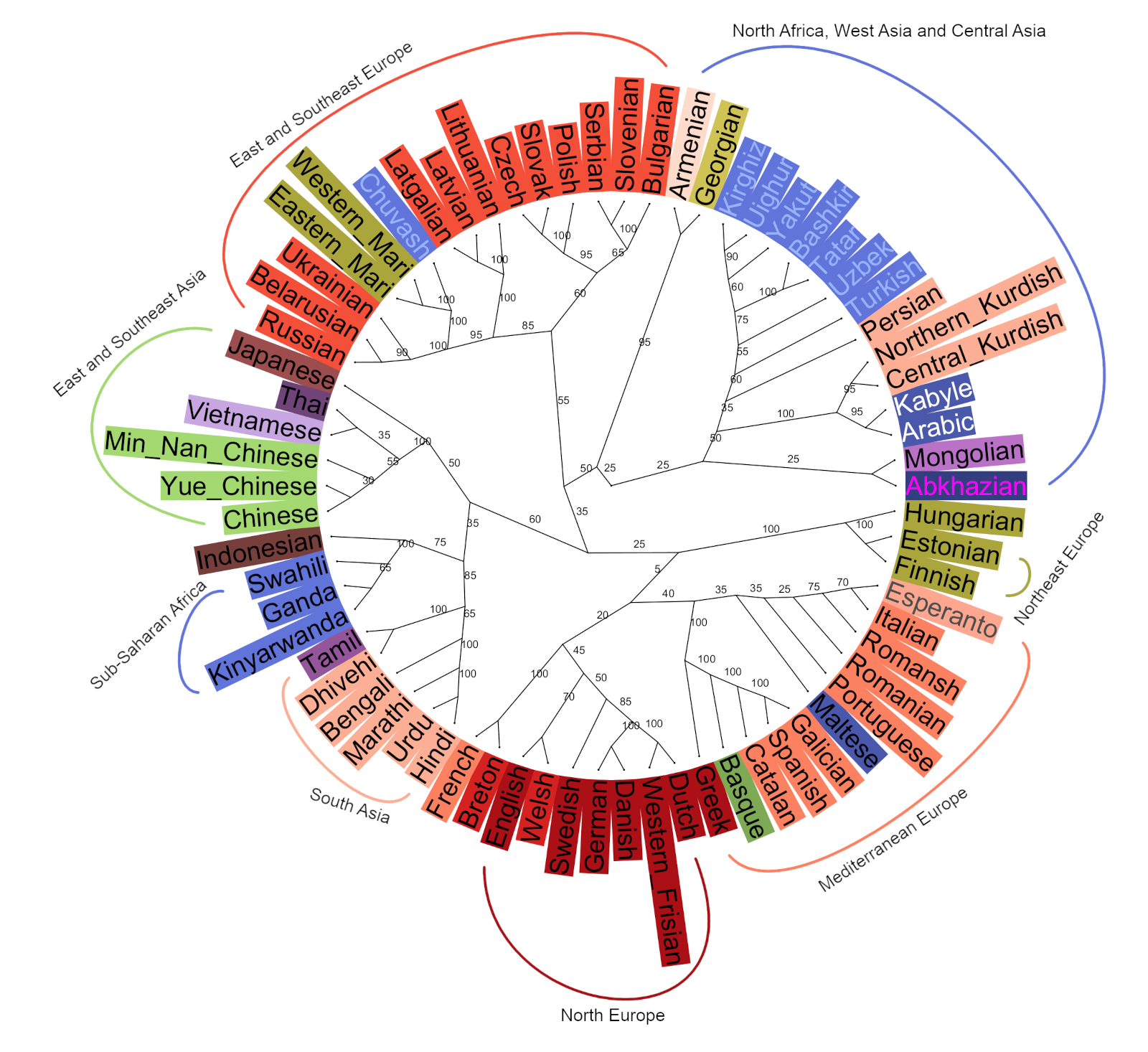}
\end{figure}

To address these questions, we tested the stability of the LDA embedding based distance measure with respect to sample size for individual languages that contain at least 1000 samples in the dataset (altogether 70 languages, see Fig~\ref{fig:Fig11}). Given a sample size $n$ (ranging from 10 to 100), we randomly selected 10 separate non-overlapping sample sets, each containing $n$ recordings for every language, and derived the language LDA embeddings, based on the $n$ samples. For each of the 10 sample sets, we generated for every language a vector of LDA embedding cosine distances between the language and every other language. Subsequently, for every language we calculated the average correlation among these 10 vectors and used it as a measure of robustness of the mutual relationship between the given language and the remaining 69 languages with respect to the particular sample size $n$. The higher the correlation, the less sample dependent is the distance-based positioning of the given language among other languages, and, \textit{vice versa}, low correlations indicate a strong impact of the choice of particular recordings on the embedding distances of the given language from other languages.

The gray line and the shaded area in Fig~\ref{fig:Fig10} depict the mean and standard deviation range of the correlations across languages as a function of sample size. For smaller sample sizes of around 10 recordings, the correlations of around 0.5 indicate relatively high dependence of embedding distances on the particular sample. For sample sizes of around 50, the correlations on average reliably exceed 0.9, showing that this amount of samples yields robust inter-language relationships regardless of the individual recordings included in the calculations. 

Fig~\ref{fig:Fig10} also contains the average correlation curves for ten outlier languages with the smallest overall sample-size based correlations. Interestingly, the list of the outlier languages largely coincides with the unconnected ``isolates'' in Figs~\ref{fig:Fig6} and \ref{fig:Fig7}. Given that the embeddings for some of these languages, e.g., Indonesian, Mongolian or Maltese, are quite distant from all other language embeddings, including language group clusters, a small change elicited by selection of different recordings may have relatively large impact on (the order of) distances from other languages and groups thereof. For some other languages, namely English, the reason for the detected instability is most likely related to a varied nature of recordings in the database including speech from different varieties of this language.

Finally, in order to evaluate the robustness of more complex family relationships among the languages using only 50 recording samples per language to calculate LDA embedding based distances, we created a consensus tree made up of 20 trees with 50 randomly distributed, non-overlapping, samples from each of the 70 languages. The resulting consensus tree is shown in Fig~\ref{fig:Fig11}, with the number representing percentage of trees having the linkage drawn in the tree. For the lower levels (more similar languages), the trees are relatively robust while relationships between large clusters are somewhat less stable. For example, the Slavic languages split to the Eastern and Western Slavic subfamilies \textit{precisely} as shown in the figure in $55 ~\%$ of the sampled trees. The Fenno-Ugric group cluster in the depicted manner--Finnish with Estonian and then subsequently with Hungarian--in all of the trees, but this group of languages cluster with the Mediterranean and North Europe languages only in $25~\%$ of trees, presumably due to somewhat ambiguous relation between Hungarian and the other languages in the dataset as indicated in Figs~\ref{fig:Fig6}, \ref{fig:Fig7} and \ref{fig:Fig10}.

\section*{Discussion}

The main insight of the presented work is that the soundscape of world languages is formed by an interaction between genetic relationships and geographical distribution of linguistic communities, relatives and neighbors. The embedding distances correlate with both genetic and geographical ones, the influence of genetic relatedness diminishing with increasing geographical distance. The clusterings depicted in Figs~\ref{fig:Fig8} and \ref{fig:Fig9} also show this interrelatedness. For example, many of the languages that do not cluster with their (sub)families are minority languages spoken in Europe; in these cases they cluster with the local majority language dominant in the very same area. This highlights the fact that language relationships are not linearly conditioned just by genealogical or geographical distance but are influenced by a number of other factors such as language contact, geopolitics and language policies.

On the local level, i.e., among genetically closely related languages, the embedding distances translated to hierarchical agglomerative trees show easily interpretable results, in many cases compatible with traditionally derived genealogical relationships, in others, geographical proximities among linguistic areas; see Appendix 2 for dendrograms for all (sub)families with at least five representatives in the dataset.

Visualizing the similarities and differences among larger language groups, and plotting a worldwide network of language relationships is somewhat more challenging. There are several practical and theoretical reasons behind the difficulties. Both the dendrograms and the NeighborNet visualization reduce the mutual relationship space to a form of two-dimensional plot based on a large number of pair-wise distances. There will inevitably be information left out when simplifying relationships into a two-dimensional space. As such, the heat map in Fig~\ref{fig:Fig5} best reflects the complicated ways the languages are interrelated.

Due to a long history of contact and separation, less related languages may share some features (e.g., phonotactics) while differing in others (e.g., rhythm). This means that similarity measures combining multiple features are not always `transitive': language A might resemble B in one feature set, and B resemble C in another, but A and C could still differ greatly. Consequently, the inclusion or exclusion of language B in datasets can significantly affect the relative positions of A and C (and their neighbors) on a global family tree, potentially altering its structure. Additionally, while A's closest relative in the dataset may be B, the dataset might include even closer relatives of B, causing A to appear distant from B in a tree constructed bottom-up.
When we, for example, remove Breton from the dataset, French clusters with Italic languages (as a separate branch) while Hebrew (linked to French and Breton in Fig~\ref{fig:Fig8}) moves next to Persian. Similarly, when we removed Albanian and Portuguese, Romanian clusters with South Slavic languages rather than within the Italic branch.

This issue is, naturally, related to the overall coverage of languages in the dataset. The CV database is currently heavily biased towards languages spoken in Europe, with relatively dense sampling of Indo-European and Dravidian languages from the Indian subcontinent, some representatives from East and Central Asia, and poor coverage of Sub-Saharan Africa. Native American, Australian and most South East Asian languages are missing. That means that while relationships among Indo-European languages are captured relatively reliably, expanding the dataset by filling the existing gaps might, in the future, reveal more complex and robust relationships on the global scale.

Keeping in mind these shortcomings, despite its relative simplicity, our methodology generates strikingly interpretable representations of language relations, even on the global scale. Outliers in the visualizations are in fact languages that lack close relatives, such as Greek and Indonesian, or are heavily influenced from various directions, such as Maltese. The dynamic nature of languages--as opposed to being fixed, abstract entities--is reflected in the results as well: though the classifier is quite reliable in predicting the correct language regardless of the utterance (most languages are predicted 100\% correct over all utterances), we still need dozens of speakers for the relationships between languages to become robust. As such, there is always some language internal variation between individual speakers that is reflected in the embedding space.

In fact, the robustness analysis shows that sound-based relationships among languages can be reliable derived from a relatively small sample-sets for individual languages: around 50 samples per language, preferably from different speakers, suffice. This opens us a realistic possibility to include progressively more and more languages in the analysis. What we need is relatively small number of samples, a handful of good (but not necessarily studio-quality) recordings by several representative, native speakers for as many languages and language communities as possible, rather than sourcing large datasets for individual languages. 

In addition to the dataset to be analyzed, the present methodology assumes an LID model can extract relevant discriminative information from the speech material from the included languages. Training such a model naturally requires a larger amount of data than those needed subsequent analyses. Importantly, our results show that it is not necessary to include all analyzed languages in the training set of the LID model: as seen in Figs~\ref{fig:Fig8} and \ref{fig:Fig9}, the languages not included in the LID model get clustered appropriately among those that are. In order to achieve reliable results for a wider range of language groups, however, the LID model would need to be adapted for a subset of languages from the areas not currently included its training materials. For these languages, a larger database is needed to perform the adaptation.

Both the discriminative learning used for training LID models and the LDA transformation might somewhat skew distances among languages by including more of detailed information useful for discriminating more similar sounding ones (`difficult cases') than for distinct pairs ('easy cases') in the embeddings. Other ML based techniques might mitigate some of these potential issues. For example, the language embeddings might be obtained from multilingual generative models such as speech synthesizers~\cite{hiovain2022comparative,suni2019comparative}.

Like most current `black-box' ML-based approaches, the present methodology does not provide direct insight on precisely which linguistic/phonetic features are behind the clustering of the languages. Given the SSL speech signal representations and LID fine-tuning, the distances between different pairs of languages can reflect multiple features, such as statistical distributions of phones, rhythmic characteristics, tonality, phrase or sentence level prosody, etc. Also, different combinations of these features might be reflected in embedding distances between different pairs of languages. In the future, the methodology may be extended to address these issues by selectively removing particular characteristics of the speech signal by, e.g., flattening pitch contours or delexicalization techniques. Changes in inter-language clustering elicited this way might provide a more detailed view on how particular characteristics of speech influence observed distances among languages. 

The current methodology has the potential to enhance existing approaches for exploring various research topics, including the histories of low-resource languages, human migration patterns, and the processes behind language change. It offers an efficient framework for analyses that does not rely on expert labor such as creating phonetic transcriptions or typological descriptions. By extracting high-dimensional continuous representations from the speech signal instead of using specific features, the presented model allows us to analyze language relationships \textit{in vivo}, minimizing researcher bias and enabling more data-driven insights.
In conjunction with our earlier work on analyzing Finnish dialectal variation~\cite{toro24_speechprosody, toro2024sociolinguistic}, our approach also shows that the embedding-based models contain information both about macro-level and micro-level variation of spoken language. In the future, the conceptual gap between language, dialect, sociolect and idiolect could be bridged by investigating dialectal and language level variation together, operationalized as distances among various speaker groups (using areal or socio-economic data) from diverse linguistic and social backgrounds.

\section*{Acknowledgments}
A large language model (ChatGPT) was used for minor stylistic editing during the writing process of this article.

%
%
%
\bibliography{plos-latex-template/references}

\begin{appendices}

\section*{Appendix 1: List of languages}
\label{app:langs}

The following table lists all the languages in the analyzed dataset, with ISO codes, language family (subfamily for Indo-European languages) and sample count.

\newpage
\begin{tabular}{l|l|l|l}
\toprule
       Language & ISO &         Language Family & No. speakers/samples  \\ 
\midrule
      Abkhazian & abk &            Abkhaz Adyge & 125 / 1000  \\
      Afrikaans & afr &                Germanic & 6 / 25  \\
       Albanian & sqi &                Albanian & 37 / 344  \\
        Amharic & amh &            Afro-Asiatic & 13 / 146  \\
         Arabic & arb &            Afro-Asiatic & 535 / 1000  \\
       Armenian & hye &                 Armenic & 150 / 1000  \\
       Assamese & asm &            Indo-Iranian & 32 / 495  \\
       Asturian & ast &                  Italic & 9 / 144  \\
    Azerbaijani & azj &                  Turkic & 10 / 26  \\
           Basa & bas &          Atlantic-Congo & 28 / 487  \\
        Bashkir & bak &                  Turkic & 385 / 1000  \\
         Basque & eus &                 Isolate & 952 / 1000  \\
     Belarusian & bel &            Balto-Slavic & 672 / 1000  \\
        Bengali & ben &            Indo-Iranian & 1000 / 1000  \\
         Breton & bre &                  Celtic & 118 / 1000  \\
      Bulgarian & bul &            Balto-Slavic & 77 / 1000  \\
        Catalan & cat &                  Italic & 962 / 1000  \\
Central Kurdish & ckb &            Indo-Iranian & 540 / 1000  \\
        Mandarin Chinese & cmn &            Sino-Tibetan & 846 / 1000  \\
        Chuvash & chv &                  Turkic & 82 / 1000  \\
          Czech & ces &            Balto-Slavic & 340 / 1000  \\
         Danish & dan &                Germanic & 103 / 1000  \\
        Dhivehi & div &            Indo-Iranian & 205 / 1000  \\
          Dutch & nld &                Germanic & 544 / 1000  \\
          Dyula & dyu &                   Mande & 11 / 47  \\
   Eastern Mari & mhr &                  Uralic & 180 / 1000  \\
        English & eng &                Germanic & 998 / 1000  \\
          Erzya & myv &                  Uralic & 10 / 431  \\
      Esperanto & epo &     Artificial Language & 470 / 1000  \\
       Estonian & ekk &                  Uralic & 343 / 1000  \\
        Finnish & fin &                  Uralic & 156 / 1000  \\
         French & fra &                  Italic & 852 / 1000  \\
       Galician & glg &                  Italic & 512 / 1000  \\
          Ganda & lug &          Atlantic-Congo & 225 / 1000  \\
       Georgian & kat &              Kartvelian & 311 / 1000  \\
         German & deu &                Germanic & 818 / 1000  \\
          Greek & ell &         Graeco-Phrygian & 229 / 1000  \\
     Hakha Chin & cnh &            Sino-Tibetan & 153 / 686  \\
          Hausa & hau &            Afro-Asiatic & 18 / 594  \\
         Hebrew & heb &            Afro-Asiatic & 10 / 217  \\
          Hindi & hin &            Indo-Iranian & 229 / 1000  \\
      Hungarian & hun &                  Uralic & 602 / 1000  \\
     Indonesian & ind &            Austronesian & 259 / 1000  \\
          Irish & gle &                  Celtic & 109 / 464  \\
        Italian & ita &                  Italic & 725 / 1000  \\
       Japanese & jpn &                 Japonic & 749 / 1000  \\
         Kabyle & kab &            Afro-Asiatic & 426 / 1000  \\
         Kazakh & kaz &                  Turkic & 97 / 452  \\
    Kinyarwanda & kin &          Atlantic-Congo & 199 / 1000  \\
        Kirghiz & kir &                  Turkic & 124 / 1000  \\
         Korean & kor &                Koreanic & 33 / 254  \\
            Lao & lao &               Tai-Kadai & 4 / 21  \\
      Latgalian & ltg &            Balto-Slavic & 134 / 1000  \\
\bottomrule
\end{tabular}

\begin{tabular}{l|l|l|l}
\toprule
         Language & ISO & Language Family & No. speakers/samples  \\
\midrule
          Latvian & lav &    Balto-Slavic & 821 / 1000  \\
         Ligurian & lij &          Italic & 8 / 575  \\
       Lithuanian & lit &    Balto-Slavic & 204 / 1000  \\
       Macedonian & mkd &    Balto-Slavic & 4 / 44  \\
        Malayalam & mal &       Dravidian & 68 / 596  \\
          Maltese & mlt &    Afro-Asiatic & 121 / 1000  \\
          Marathi & mar &    Indo-Iranian & 49 / 1000  \\
  Min Nan Chinese & nan &    Sino-Tibetan & 101 / 1000  \\
           Moksha & mdf &          Uralic & 8 / 95  \\
        Mongolian & mon & Mongolic Khitan & 268 / 1000  \\
           Nepali & nep &    Indo-Iranian & 15 / 177  \\
 Northern Kurdish & kmr &    Indo-Iranian & 290 / 1000  \\
Norwegian Nynorsk & nno &        Germanic & 26 / 290  \\
          Occitan & oci &          Italic & 67 / 203  \\
            Oriya & ory &    Indo-Iranian & 36 / 626  \\
         Ossetian & oss &    Indo-Iranian & 5 / 32  \\
          Panjabi & pan &    Indo-Iranian & 40 / 437  \\
          Persian & pes &    Indo-Iranian & 667 / 1000  \\
           Polish & pol &    Balto-Slavic & 648 / 1000  \\
       Portuguese & por &          Italic & 639 / 1000  \\
           Pushto & pus &    Indo-Iranian & 7 / 177  \\
         Romanian & ron &          Italic & 231 / 1000  \\
          Romansh & roh &          Italic & 79 / 1000  \\
          Russian & rus &    Balto-Slavic & 634 / 1000  \\
          Santali & sat &   Austroasiatic & 8 / 131  \\
          Saraiki & skr &    Indo-Iranian & 40 / 902  \\
        Sardinian & sro &          Italic & 9 / 200  \\
          Serbian & srp &    Balto-Slavic & 120 / 1000  \\
           Slovak & slk &    Balto-Slavic & 140 / 1000  \\
        Slovenian & slv &    Balto-Slavic & 119 / 1000  \\
          Spanish & spa &          Italic & 972 / 1000  \\
          Swahili & swh &  Atlantic-Congo & 463 / 1000  \\
          Swedish & swe &        Germanic & 443 / 1000  \\
        Tamazight & zgh &    Afro-Asiatic & 3 / 38  \\
            Tamil & tam &       Dravidian & 356 / 1000  \\
            Tatar & tat &          Turkic & 187 / 1000  \\
           Telugu & tel &       Dravidian & 7 / 26  \\
             Thai & tha &       Tai-Kadai & 890 / 1000  \\
            Tigre & tig &    Afro-Asiatic & 4 / 39  \\
          Turkish & tur &          Turkic & 561 / 1000  \\
          Turkmen & tuk &          Turkic & 45 / 436  \\
           Uighur & uig &          Turkic & 466 / 1000  \\
        Ukrainian & ukr &    Balto-Slavic & 454 / 1000  \\
    Upper Sorbian & hsb &    Balto-Slavic & 17 / 398  \\
             Urdu & urd &    Indo-Iranian & 171 / 1000  \\
            Uzbek & uzb &          Turkic & 629 / 1000  \\
       Vietnamese & vie &   Austroasiatic & 133 / 1000  \\
            Welsh & cym &          Celtic & 550 / 1000  \\
  Western Frisian & fry &        Germanic & 625 / 1000  \\
     Western Mari & mrj &          Uralic & 38 / 1000  \\
            Yakut & sah &          Turkic & 80 / 1000  \\
           Yoruba & yor &  Atlantic-Congo & 87 / 898  \\
      Yue Chinese & yue &    Sino-Tibetan & 697 / 1000  \\
\bottomrule
\end{tabular}

\section*{Appendix 2: Dendrograms for related languages}
\label{app:dendrograms}

\renewcommand{\thefigure}{S\arabic{figure}}
\setcounter{figure}{0}

\begin{figure*}[!ht]
\begin{adjustwidth}{-2.25in}{0in} 

\caption{{\bf Dendrogram of Germanic languages.}}
\label{fig:germanic}
\includegraphics[width=1.\linewidth]{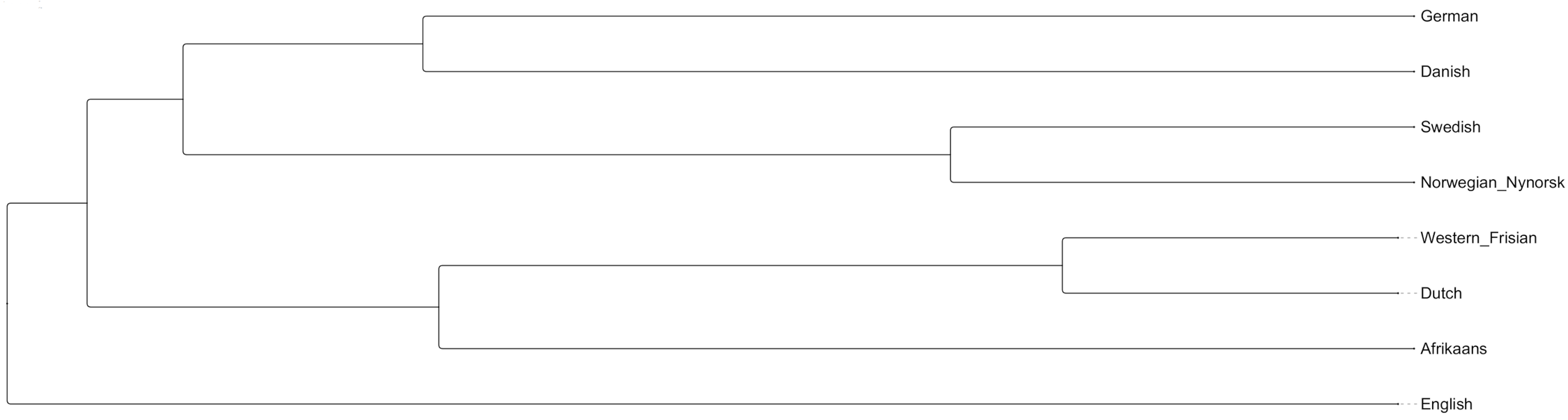}
\end{adjustwidth}
\end{figure*}

\begin{figure*}[!ht]
\begin{adjustwidth}{-2.25in}{0in} 

\caption{{\bf Dendrogram of Uralic languages.}}
\label{fig:uralic}
\includegraphics[width=1.\linewidth]{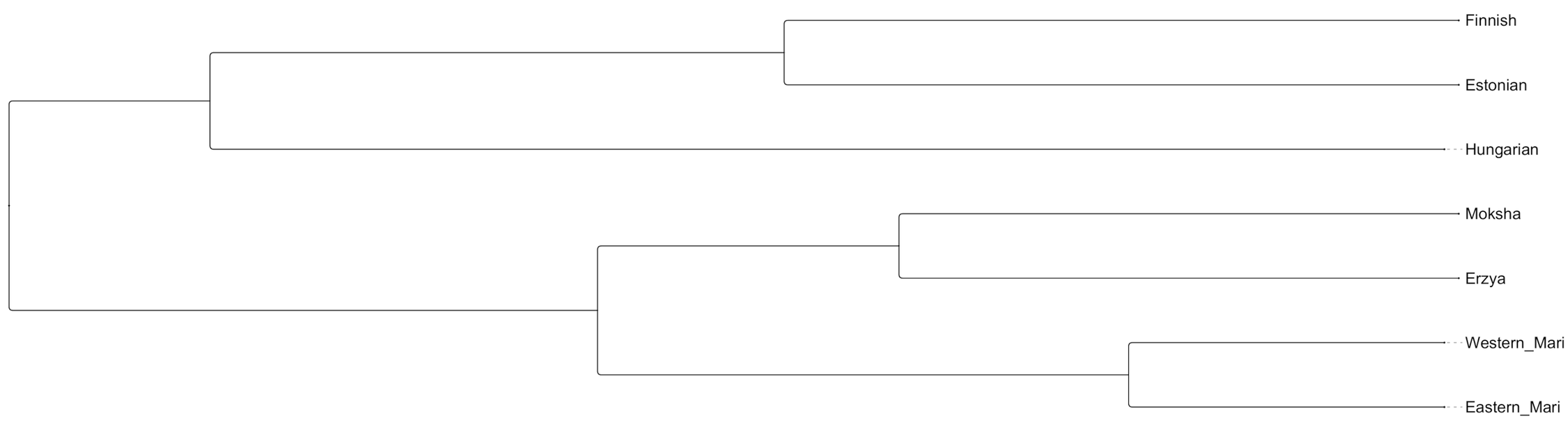}
\end{adjustwidth}
\end{figure*}

\begin{figure*}[!ht]
\begin{adjustwidth}{-2.25in}{0in} 

\caption{{\bf Dendrogram of Turkic languages.}}
\label{fig:turkic}
\includegraphics[width=1.\linewidth]{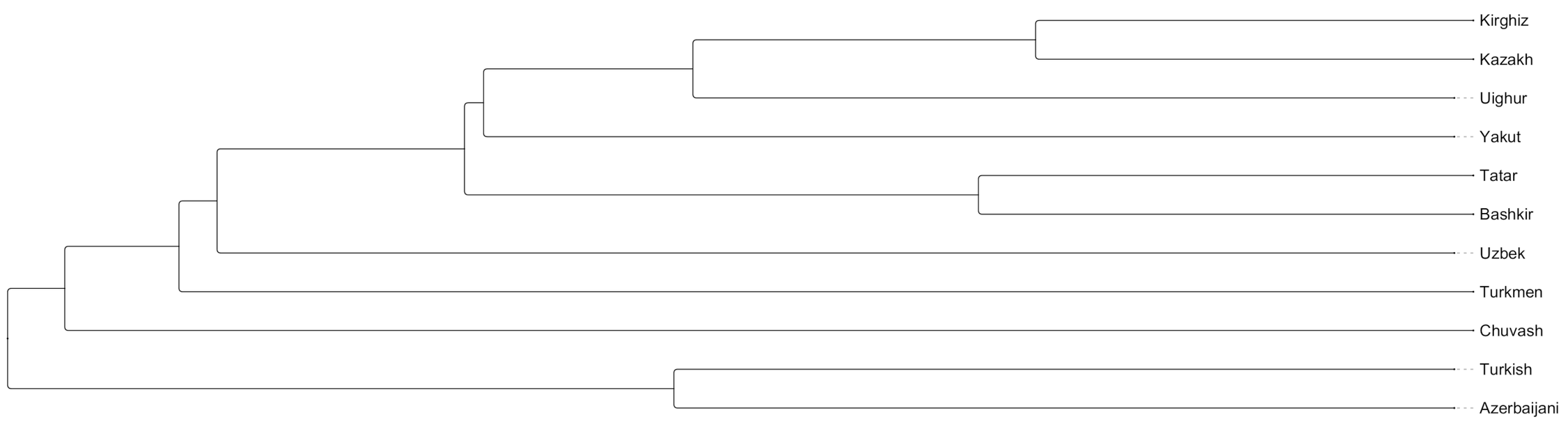}
\end{adjustwidth}
\end{figure*}

\begin{figure*}[!ht]
\begin{adjustwidth}{-2.25in}{0in} 

\caption{{\bf Dendrogram of Indo-Iranian languages.}}
\label{fig:indo}
\includegraphics[width=1.\linewidth]{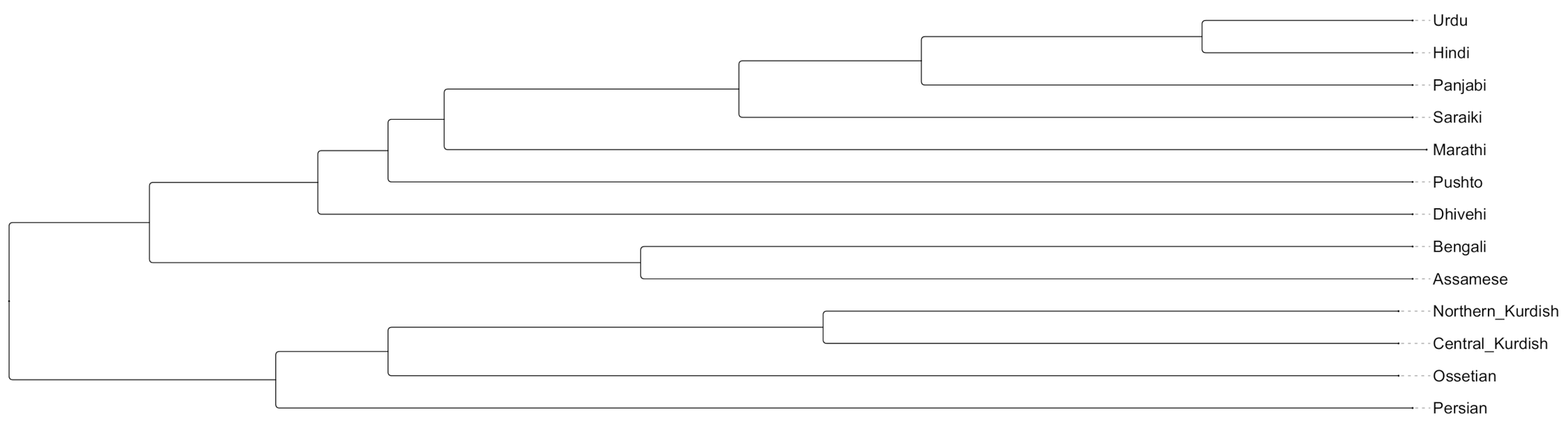}
\end{adjustwidth}
\end{figure*}

\begin{figure*}[!ht]
\begin{adjustwidth}{-2.25in}{0in} 

\caption{{\bf Dendrogram of Italic languages.}}
\label{fig:italic}
\includegraphics[width=1.\linewidth]{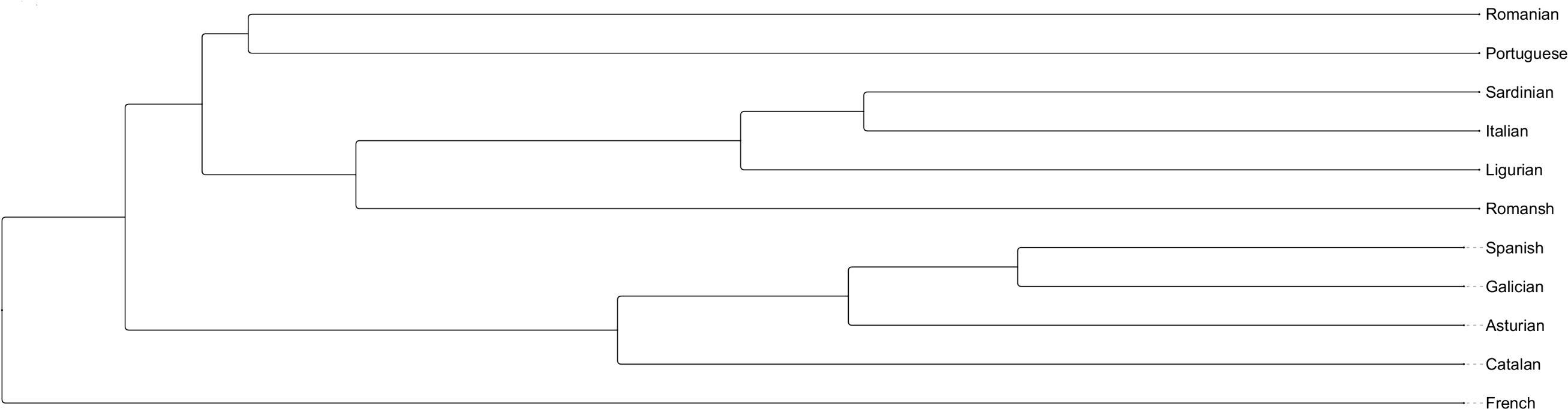}
\end{adjustwidth}
\end{figure*}

\begin{figure*}[!ht]
\begin{adjustwidth}{-2.25in}{0in} 

\caption{{\bf Dendrogram of Slavic languages.}}
\label{fig:slavic}
\includegraphics[width=1.\linewidth]{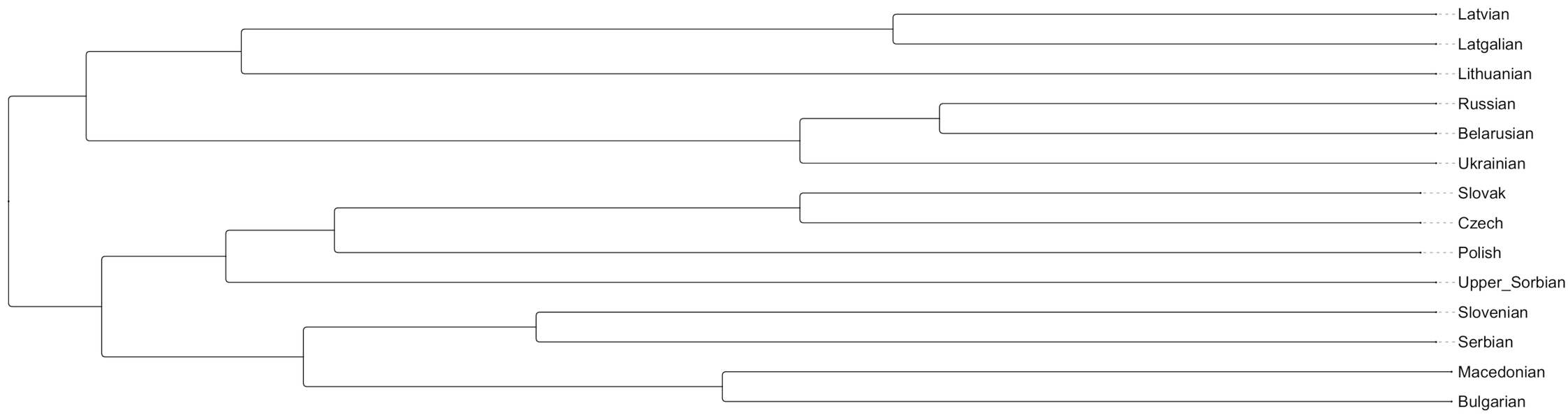}
\end{adjustwidth}
\end{figure*}

\begin{figure*}[!ht]
\begin{adjustwidth}{-2.25in}{0in} 

\caption{{\bf Dendrogram of Afro-Asiatic languages.}}
\label{fig:afroasiatic}
\includegraphics[width=1.\linewidth]{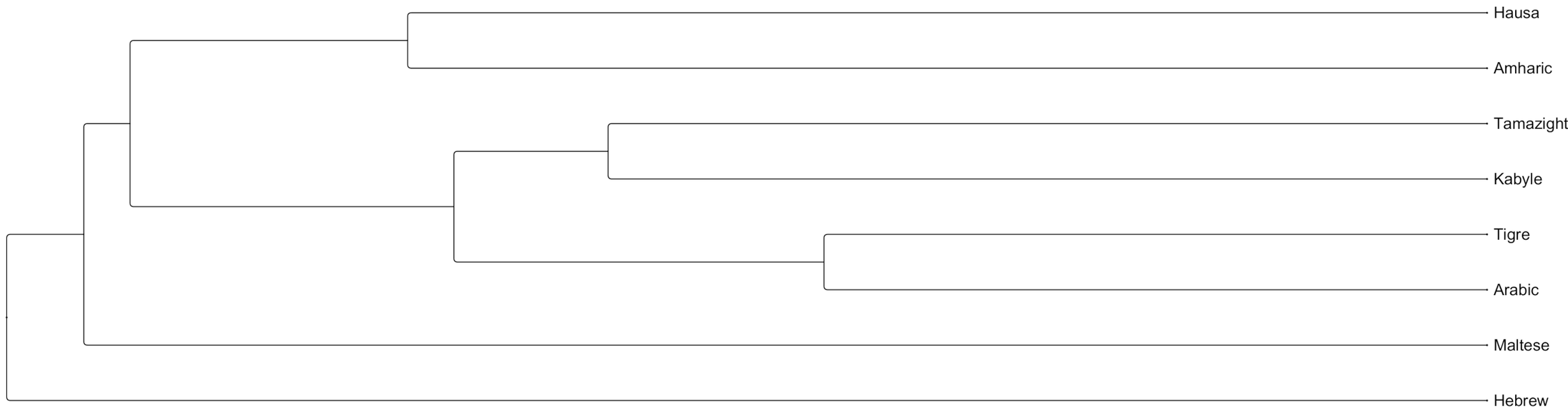}
\end{adjustwidth}
\end{figure*}

\end{appendices}

\end{document}